\documentclass{elsarticle}
\usepackage{amssymb}
\journal{Information Fusion}
\usepackage{setspace}
\usepackage{graphicx}
\usepackage{textcomp}
\usepackage{float}
\usepackage[table,xcdraw]{xcolor}
\usepackage{CJKutf8}
\usepackage{setspace}
\usepackage{booktabs}
\usepackage{amsfonts}
\usepackage{amsthm,amsmath}
\usepackage{mathrsfs}
\usepackage{multirow}
\usepackage[noend]{algpseudocode}
\usepackage{algpseudocode}
\usepackage{diagbox}
\usepackage{color}
\usepackage{colortbl}  
\usepackage{xcolor}
\usepackage{array} 
\usepackage{bbding} 
\usepackage{graphicx}
\usepackage[marginal]{footmisc}

\begin{document}
	
	\begin{frontmatter}
		
		\author[1]{Zhimeng Xin}
		

		\author[2]{Shiming Chen}

		
		\author[2]{Tianxu Wu}
		
		\author[2]{Yuanjie Shao}
		
		\author[3]{Weiping Ding \corref{cor1}}	
		
		
		\author[1]{Xinge You \corref{cor1}}
	

		\cortext[cor1]{Corresponding author: youxg@mail.hust.edu.cn (Xinge You) and dwp9988@163.com (Weiping Ding)}
		
		
		\address[1]{School of Cyber Science and Engineering, Huazhong University of Science and Technology, Wuhan {\rm 430074}, China}
		
		
		\address[2]{School of Electronic Information and Communications, Huazhong University of Science and Technology, Wuhan {\rm 430074}, China}
		
		\address[3]{School of Information Science and Technology, Nantong University, Nantong {\rm 226019}, China }

		\title{Few-Shot Object Detection: Research Advances and Challenges}

		\begin{abstract}

			Object detection as a subfield within computer vision has achieved remarkable progress, which aims to accurately identify and locate a specific object from images or videos. Such methods rely on large-scale labeled training samples for each object category to ensure accurate detection, but obtaining extensive annotated data is a labor-intensive and expensive process in many real-world scenarios. To tackle this challenge, researchers have explored few-shot object detection (FSOD) that combines few-shot learning and object detection techniques to rapidly adapt to novel objects with limited annotated samples. This paper presents a comprehensive survey to review the significant advancements in the field of FSOD in recent years and summarize the existing challenges and solutions. Specifically, we first introduce the background and definition of FSOD to emphasize potential value in advancing the field of computer vision. We then propose a novel FSOD taxonomy method and survey the plentifully remarkable FSOD algorithms based on this fact to report a comprehensive overview that facilitates a deeper understanding of the FSOD problem and the development of innovative solutions. Finally, we discuss the advantages and limitations of these algorithms to summarize the challenges, potential research direction, and development trend of object detection in the data scarcity scenario.

		\end{abstract}


	\end{frontmatter}

	\section{Introduction}
	
	In recent years, object detection algorithms have made remarkable progress, which aims to accurately identify and locate a specific object from an image or video \cite{ifob1,ifob2,ifob3}. However, generic object detection methods usually require a large number of annotated samples to train the model, which is often time-consuming, impractical, or expensive work in practical applications \cite{TFA,metafrcn}. Therefore, detecting objects in the data-scarcity scenarios is an important way to solve the above problems \cite{lstd}.
	
	Fortunately, few-shot learning (FSL) researchers \cite{fsl1,fsl2,fslsurvey} found that even the children who had already learned the knowledge of a dog could learn the concept of a wolf with only a few samples. It is assumed that the machine model also can utilize prior knowledge to rapidly learn the few-shot categories in the same way as humans. Based on extensive experiments, FSL researchers found that training the model under the premise of a large amount of prior knowledge can adapt to the data with only a few samples. The machine model can appropriately migrate and generalize in the data-scarcity and unseen scenarios \cite{fsl3,iffler1,iffler2}. Therefore, combining FSL with object detection for \textbf{f}ew-\textbf{s}hot \textbf{o}bject \textbf{d}etection (\textbf{FSOD}) is a promising research field, which enables the model to quickly adapt to a few-shot number of annotated objects without weakening the performance \cite{acmfsod1,acmfsod2}.
	
\begin{figure}[t]
		\vspace{0.1cm}
		\centering
		\includegraphics[width=9.68cm,height=7.38cm]{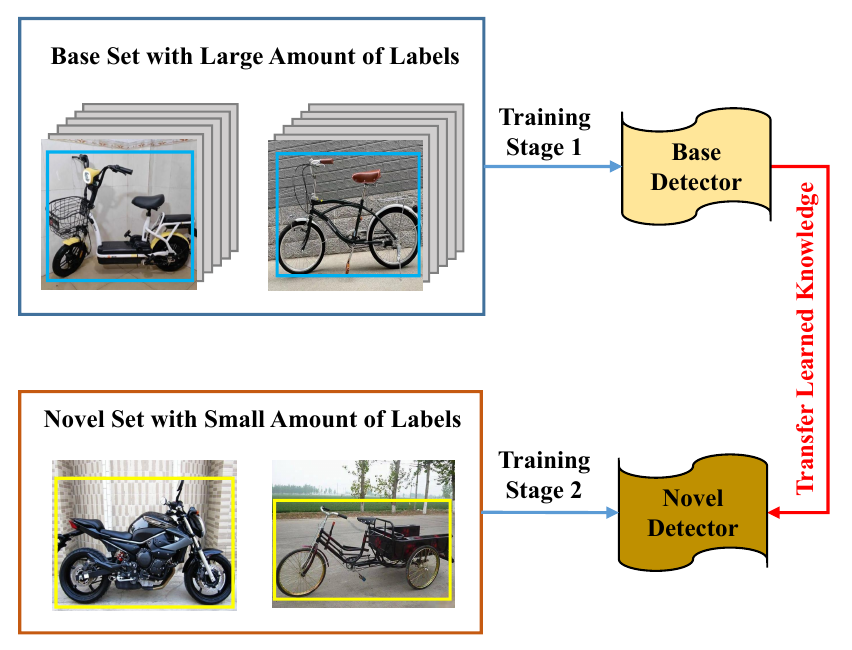} 
		\vspace{-0.3cm}
		
		\caption{
			Processing of mainstream FSOD.
		}

		\label{f1}
		\vspace{-0.1cm}
	\end{figure}

	\textcolor{blue}{At present, mainstream FSOD methods borrow ideas from few-shot learning to train the detection network in the data-scarcity scenarios with the help of the prior knowledge learned in the well-annotated base class \cite{lstd,metadet}. The processing of such FSOD concepts is derived from transfer learning that leverages knowledge or experience gained from one task to improve the performance of another related task. Figure \ref{f1} as an example illustrates the processing of most current FSOD methods that refer to the training of a base class detection model on a base dataset with a large amount of annotation information, and the detection of a novel target is realized only by using the novel set with few annotations and the prior knowledge provided by the base class model. Based on the above analysis, we consider that all FSOD models that are trained from the base to the novel stage follow the concept of transfer learning.}

	\textcolor{blue}{	\textbf{Taxonomy.} According to the idea of transfer learning, we thus classify the FSOD methods based on the above two-stage training strategy into two broad categories, including \textbf{episode-task-based} and \textbf{single-task-based methods}. Specifically, episode-task-based methods, following the principle of meta-learning ideas \cite{metadet,metal1,metal2}, divide the detection task into a series of episode tasks with few-shot samples in base and novel stages, which assists the model in rapidly adapting to the detection task in the data-scarcity scenario. Instead of repeated episode tasks, single-task-based methods directly transfer the original or fine-tuned parameters of the base model to the novel stage and then fine-tune the few-shot detection task in the novel stage \cite{cov}. By straightforwardly utilizing the parameters learned from the source domain model, FSOD models of two classifications can quickly suitable for the unseen objects in the target domain and improve their performance using the limited annotated samples available. Based on this proposed method taxonomy, we survey the plentifully remarkable FSOD algorithms to report a comprehensive overview that facilitates a deeper understanding of the FSOD problem and the development of innovative solutions. }
	
	\textcolor{blue}{	\textbf{Comparison with Related Surveys.} There are some related surveys \cite{survey1,survey2,survey3,survey4,survey5,survey6,survey7} for FSOD. Compared with them, we cover the latest FSOD research and propose a novel taxonomy to provide a more comprehensive FSOD survey. The superiorities of our survey can be summarized as follows:\textbf{ (1) Latest FSOD surveys.} These surveys \cite{survey1,survey5,survey6} only research the early FSOD works, the latest FSOD methods, e.g., the FSL combined with the recently popular transformer-based framework fail to report. \cite{survey4} is not a specific FSOD review, which neglects some critical problems. For example, catastrophic forgetting problems of base classes in generalized FSOD \cite{cvpr2023gfsod} are not mentioned in \cite{survey4}. Compared with them, we report the latest FSOD algorithms and introduce the methods and their results of keeping the performance of base objects in generalized FSOD. \textbf{(2) Clear Taxonomy.} Recent surveys \cite{survey2,survey3,survey7} consider that the FSOD methods based on two-stage training that adopt meta-learning to quickly adapt novel tasks are independent of transfer learning, which may be inaccurate according to the transfer learning concept. We consider that all FSOD models that are trained from the base to the novel stage follow the idea of transfer learning. Therefore, in our survey, such meta-learning-based FSOD algorithms are classified as special transfer learning methods. \textbf{(3) Exploring problems and solutions in FSOD.} Compared with recent surveys, we also focus on the important problems found in FSOD by current FSOD methods and how they propose corresponding solutions based on these problems. We consider that this can not only let readers understand the development of methods in the field of FSOD but also understand the major problems that have been solved in this field, which can assist the follow-up researchers to keep studying the unsolved problems. }
	
	In view of this, the main contribution of our survey can be summarized as follows:
	
	\begin{itemize}

		\item We first formulate a novel taxonomy scheme for the existing FSOD methods that consist of episode-task-based and single-task-based research according to the definition of transfer learning.

		\item We then survey the plentifully remarkable FSOD algorithms based on our proposed method classification to report a comprehensive overview that facilitates a deeper understanding of the FSOD problem and the development of innovative solutions.

		\item We finally discuss the advantages and limitations of such FSOD algorithms to summarize the challenges, potential research direction, and development trend of object detection in the data scarcity scenario.

	\end{itemize}

	The remainder of this article is organized as follows: Section 2 is the problem setting and benchmark datasets of FSOD approaches. We then develop a novel taxonomy of FSOD methods by the definition of transfer learning in Section 3. Furthermore, we comprehensively introduce the motivations and solutions of the two types of FSOD methods in Sections 4 and 5, respectively, and highlight the differences as well as similarities in Section 6. Finally, Sections 7 and 8 separately provide the challenges, developing trends, and application fields of FSOD and the conclusion of our survey.

	\section{Formulation of FSOD}
	
	\subsection{Problem Setting}
	
	Currently, the training process of most FSOD methods is based on two stages \cite{TFA,acmfsod1,defrcn}. First, a base model $M_{base}$ is trained on a large-scale dataset $D_{base}$, where all target classes in $D_{base}$ are denoted as $C_{base}$. Then, the prior knowledge learned in the base model is transferred to a few-shot dataset $D_{novel}$ with few annotations, where all categories in $D_{novel}$ are denoted as $C_{novel}$ and $C_{base} \cap C_{novel}=\emptyset$. \textcolor{blue}{ Given a sample $ (x’, y’) \in D_{base} \cup D_{novel}$, in which $ x’ $ is the input image with $ n $ objects with its annotation $y=\left\{ (c_i, b_i), i = 1, ..., n \right\} $, including the classification $ c_i \in C_{base} \cup C_{novel}$ and box $ b_i $ of the $ c_i $ localization corresponding to the $x’$.}
	
	On the other hand, when the novel phase trains a balanced novel dataset $D_{balance}$, the category $C_{balance} = C_{novel} \cup C_{base}$, and the number of $C_{base}$ is equal to $C_{novel}$ during training. Training a detection network with this balanced dataset is known as generalized few-shot object detection (G-FSOD) \cite{cfa, retina}. Compared with only training $C_{novel}$, G-FSOD is a more comprehensive and balanced detection approach by considering both base and novel classes, addressing the class imbalance, and evaluating the model's performance on a unified test set. G-FSOD allows for a more generalized and fair assessment of the model's ability to detect objects across different classes, including both known and unknown categories.
	
	Denote $M_{init}$ as the initialized few-shot detection model, then the final model of FSOD and G-FSOD can be given by
	
	\begin{equation}				
	\mathcal{M}_{\text {init }} \stackrel{D_{base}}{\Longrightarrow} \mathcal{M}_{\text {base }} \stackrel{D_{novel} / D_{balance}}{\Longrightarrow} \mathcal{M}_{novel} ,
	\end{equation}	
	where $ \Longrightarrow $ represents the training process.	
	
	On the other hand, Faster R-CNN \cite{fasterrcnn} is a widely used two-stage object detection architecture that combines backbone, region proposal network (RPN), and region-of-interests (RoI) for accurate and efficient object detection. Meanwhile, this detector is favored by FSOD researchers and has been successfully applied in detection with few-shot scenarios.
	
	\subsection{Benchmark Datasets and Evaluation Protocols}
	
	There are two mainly benchmarked datasets, i.e., Pascal VOC \cite{voc} and MS COCO \cite{coco} datasets to evaluate the performance of existing FSOD methods. In addition, some articles \cite{lstd,fan,TFA} have chosen to evaluate their FSOD methods on FSOD \cite{fan}, ImageNet-LOC \cite{imagenet}, or LVIS \cite{lvis} datasets.	
	
	\subsubsection{Pascal VOC Dataset}
	
	The Pascal VOC (Visual Object Classes) dataset \cite{voc} is a popular benchmark dataset for object detection and semantic segmentation. This dataset needs to define specific protocols and procedures for adapted FSOD. However, it is worth noting that the Pascal VOC dataset does not have explicit annotations or splits for FSOD. Researchers have developed methodologies to adapt it for few-shot scenarios \cite{TFA,metafrcn,metarcnn}. Based on this fact, the categories in this dataset are divided into a simple division, in which 15 categories are used to train the base model and 5 categories to train the novel model, i.e., $C_{base}=15$  and $C_{novel}=5$. Finally, the Pascal VOC dataset is split into three sets, where split Set 1 = (bird, bus, cow, motorbike, sofa), Set 2 = (aeroplane, bottle, cow, horse, sofa), and Set 3 = (boat, cat, motorbike, sheep, sofa).
	
	\subsubsection{Evaluation Protocols on the Pascal VOC Dataset}
	
	The average precision (AP) metric \cite{coco} evaluation is utilized for the Pascal VOC dataset \cite{voc}, which is calculated from the Precision-Recall (PR) curve. The value of AP is described as follows.
	
	Precision and recall calculate the object localization that is predicted box by the detector. The quality of the prediction box is determined by Intersection over Union (IoU) \cite{IOU}. IoU is calculated by dividing the area of intersection by the area of union between the predicted and ground truth (GT) bounding boxes to measure the overlap between predicted bounding boxes (PB) and ground truth bounding boxes, i.e., IoU = (PB $\cap$ GT)/( PB $\cup$ GT).
	
	Precision is the ratio of true positive (TP) detections to the total number of positive detections (TP + false positive (FP)), i.e., Precision = TP / (TP + FP). Recall is the ratio of TP detections to the total number of ground truth positive instances (TP + false negative (FN)), i.e., Recall = TP / (TP + FN). Different recall and precision values can be obtained by changing the confidence threshold or IoU threshold when calculating the PR curve. Each threshold corresponds to a point and all points are connected to form a curve. Then, AP is computed by taking the area under the PR curve. In fact, the area under the PR curve is usually estimated by discretization in the FSOD task and thus AP can be given by
	\begin{equation}
	AP = \frac{1} {\mathbb{N}} \sum_{i=1}^ \mathbb{N} P_i \cdot \Delta R_i 
	\end{equation}
	where, $ \mathbb{N} $ represents the total number of recall or precision prediction, $ P_i $ and $ R_i $ are precision and recall at $ i $th prediction. In the Pascal VOC evaluation, bAP50 (matching threshold is 0.5) evaluates the training effect of the base classes and nAP50 evaluates the performance of the entire novel objects \cite{TFA}. 
	

	\subsubsection{MS COCO Dataset}
	
	Similar to Pascal VOC, the MS COCO dataset \cite{coco} is primarily used for standard object detection and needs to define specific protocols and procedures for adapted FSOD. Instead of VOC, COCO is split once for FSOD. Specifically, there are 80 classes in the MS COCO dataset, of which 20 classes same as VOC classes are divided into novel classes and the remaining 60 classes are base classes. 
	
	\subsubsection{Evaluation Protocols on the MS COCO Dataset}
	
	Compared with VOC, COCO adopts more AP metrics \cite{coco}. Here, AP uses 10 IoU thresholds of .50:.05:.95 as the primary challenge metric in COCO. AP75 uses an IoU threshold of .75 as a strict metric. APS, APM, and APL verify the small (area $\leq$ 32*32), medium (32*32 $\leq$ area $\leq$ 96*96), and large (area $\leq$ 96*96) objects, respectively, where area represents the object size.
	
	\subsubsection{LVIS, FSOD, and ImageNet Datasets}	
	
	The LVIS dataset \cite{lvis} is a large-scale dataset for instance segmentation and object detection tasks. It provides extensive annotations for a wide range of object categories, allowing for diverse training and evaluation. In \cite{TFA}, authors define specific protocols for FSOD model training in this dataset. The evaluation criteria are the same as COCO \cite{coco}. Furthermore, the FSOD dataset \cite{fan} is specifically built to detect few-shot objects. This dataset with 1000 categories is derived from ImageNet and Open Image. Fan \textit{et al.} \cite{fan} choose 200 classes to train the novel model and the remaining 800 to the base model. In fact, training the FSOD model on LVIS or FSOD datasets is challenging due to the diversity of its categories and the large variance in the aspect ratio of images. Therefore, rare researchers use them to evaluate their proposed FSOD methods. In addition, currently, the ImageNet dataset \cite{imagenet} is primarily used to train a pre-trained model. 

	\begin{figure}[t]
		\vspace{0.1cm}
		\centering
		\includegraphics[width=12cm,height=4.82cm]{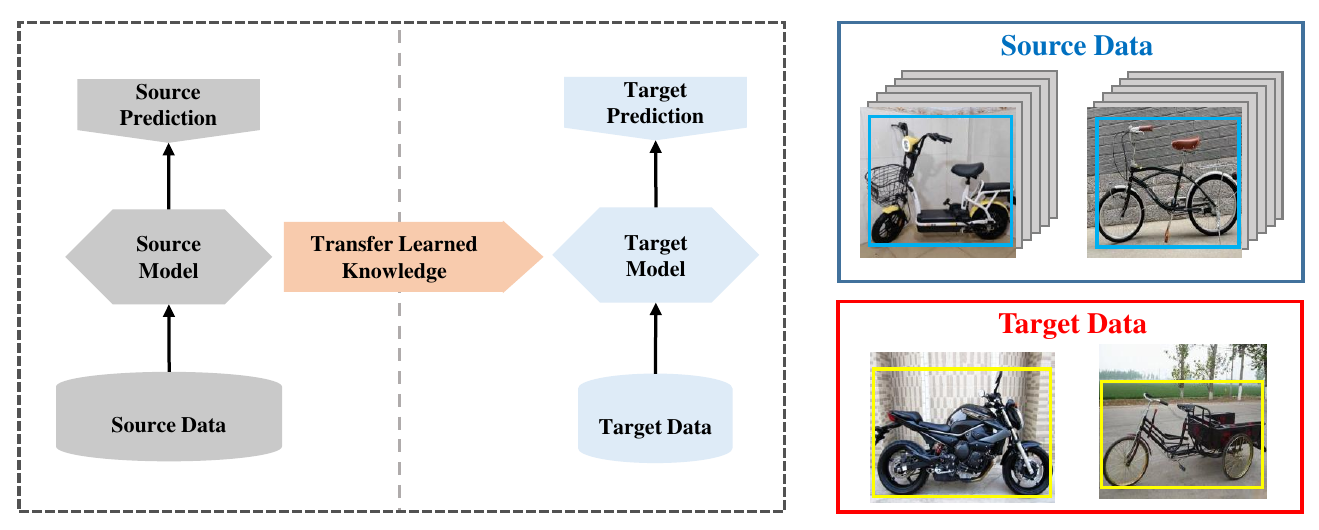} 
		\vspace{-0.3cm}
		
		\caption{
			The processing of transfer learning.
		}

		\label{transfer}
		\vspace{-0.1cm}
	\end{figure}

	\section{Taxonomy of FSOD Methods}
	
	\textcolor{blue}{In this section, we classify the current FSOD methods based on the two-stage training strategy into episode-task-based and single-task-based methods by revisiting the definition of transfer learning. We then illustrate the operation details of the two groups' methods, respectively.}
	
	\subsection{Definition of Transfer Learning}
	
	Transfer learning \cite{trans1,trans2,trans3} as an inspiring technique has been successfully used on deep models and involves leveraging knowledge gained from one task or domain to improve performance on another related task or domain.  Figure \ref{transfer} as an example illustrates the processing of transfer learning. It is like a person can quickly learn to drive a motorcycle when he learns to ride a bicycle or an electric bike. \textcolor{blue}{ Denote $D_s$ and $T_s$ as the source domain with its target domain $D_t$ and learning task with its target task $T_t$, respectively. The purpose of transfer learning is to leverage the knowledge learned from a source domain $D_s$ using a source task $D_t$, and apply this acquired knowledge to improve the prediction performance of a target task $T_t$ in a target domain $T_s$, where $D_s \neq T_s $ or $D_t \neq T_t$ \cite{trans1}.}
	
\begin{figure}[t]
		\vspace{0.1cm}
		\centering
		\includegraphics[width=11.7cm,height=7.9cm]{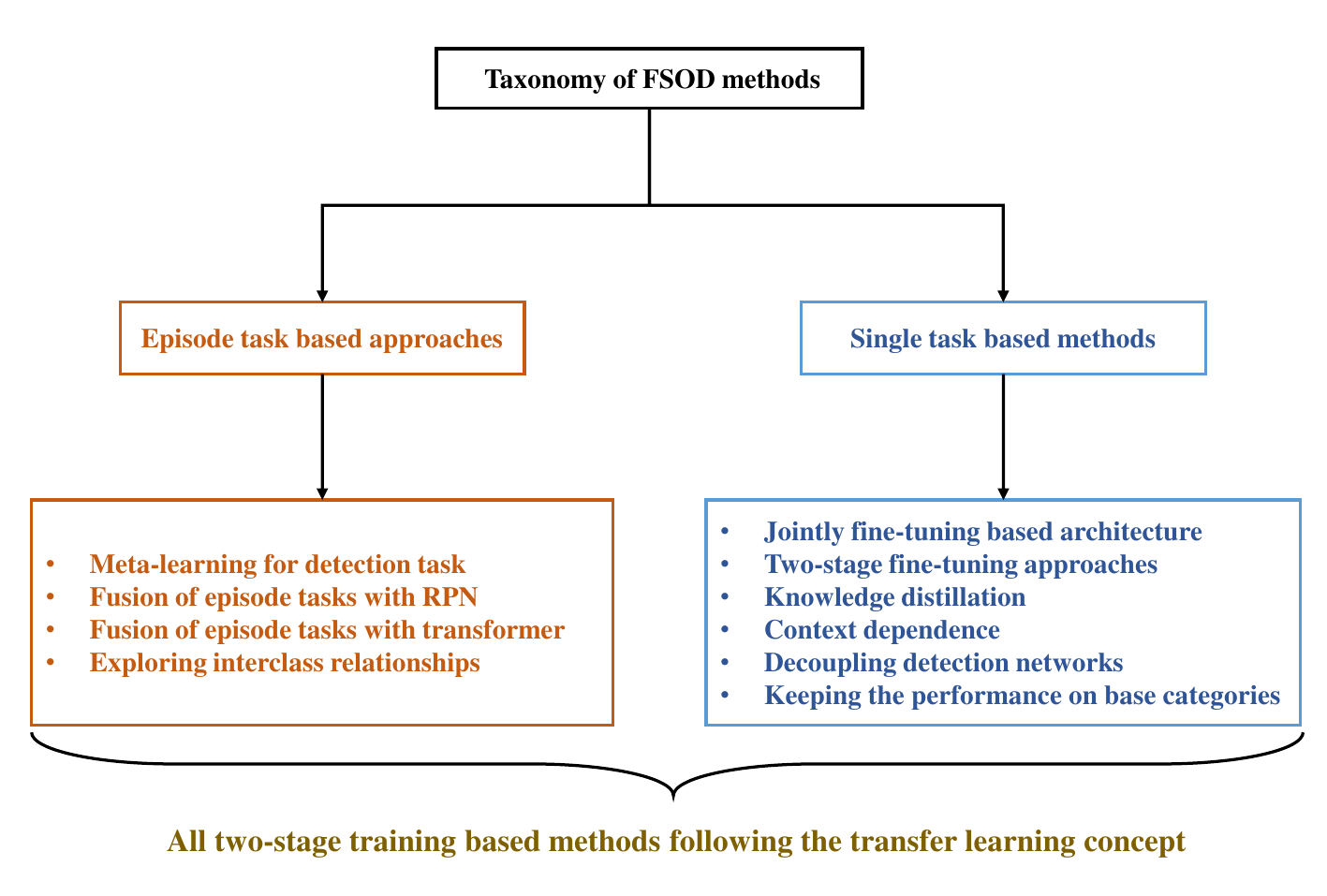} 
		\vspace{-0.3cm}
		
		\caption{
				The organizational structure of FSOD taxonomy in this survey.
		}

		\label{context}
		\vspace{-0.1cm}
	\end{figure}

	
	\textcolor{blue}{In FSOD, a pre-trained model $D_t$ that has been trained on a large-scale dataset $D_s$ is used as a starting point for a new task $T_t$ with limited labeled data $T_s$, where $D_s \neq T_s $ and $D_t = T_t$. Therefore, we consider that all FSOD methods based on the two-stage training follow transfer learning, and we then classify them into episode-task-based and single-task-based methods on this definition. Figure \ref{context} shows the organizational structure of FSOD taxonomy in our survey. In addition, this figure also illustrates the follow-up organization of this classification.}

	\begin{figure}[t]
		\vspace{0.1cm}
		\centering
		\includegraphics[width=7.7cm,height=7cm]{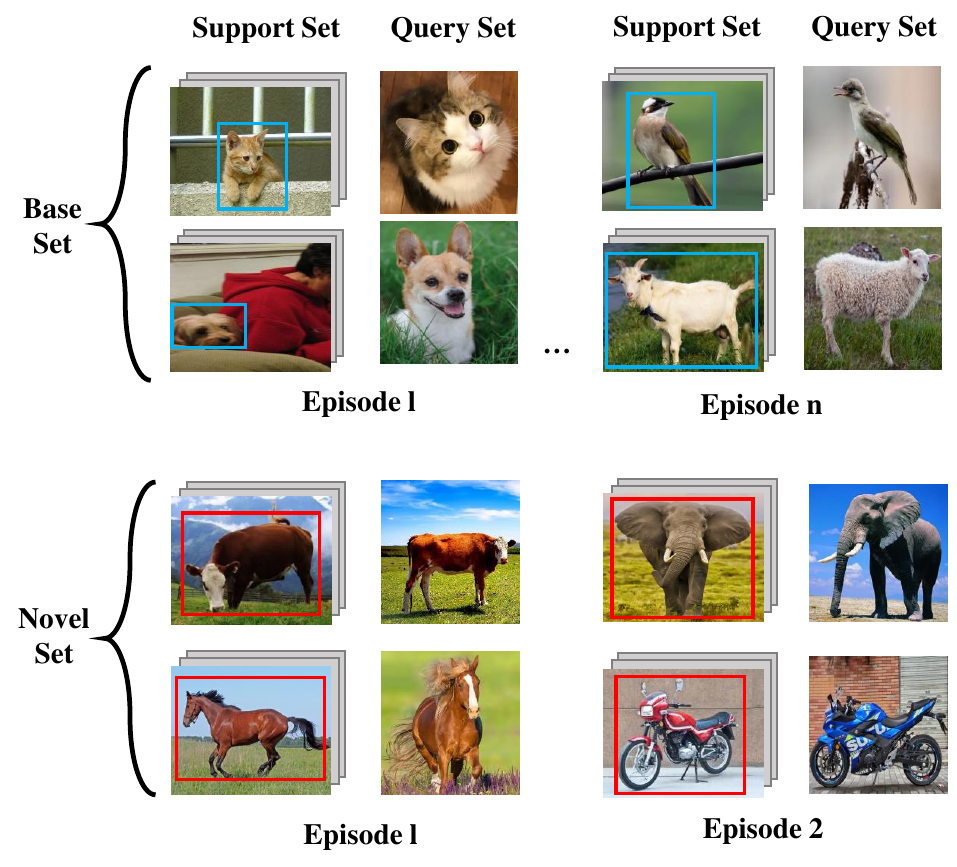} 
		\vspace{-0.3cm}
		
		\caption{
			The processing of episode-task-based approach.
		}

		\label{fm1}
		\vspace{-0.1cm}
	\end{figure}
	
	\subsection{Episode Task Based Approaches}
	
	
	\textcolor{blue}{	FSOD based on the episode approach draws inspiration from the concept of meta-learning \cite{metal1,metal2,metal3,metal4,tip2}. In the meta-learning-based approach \cite{fsl1,fsl2,fslsurvey}, the task is organized into a series of episodes, where each episode simulates an FSOD scenario. Each episode consists of the few-shot training (or support) and test (or query) set, in which, the support set is used for model training on a limited number of samples, and the query set is used to evaluate the detection performance of the model on novel objects. Denote $S$ represents the support set. Each task in the $S$ consists of $N$ classes and $K$ images per class. An episode task can also be called \textbf{\textit{N-way K-shot}} detection \cite{cvpr2022sylp}. The support set can be represented as a collection of tasks: $S=\left\{(X_i,Y_i)\right\}$, where $(X_i,Y_i)$ is the set of input images and corresponding annotations for $i$th task. Denote $Q=\left\{(X_j)\right\}$ as the query set that includes a separate collection of tasks, each with input images $X_j$ but without annotations.}
	
	The reason why $S$ typically contains only a few annotated samples is to reflect the real-world scenario where only a few samples are used to learn the features of novel objects \cite{metadet}. 	\textcolor{blue}{Therefore, by iterative training and evaluating the model on a series of episodes, the model can learn from limited samples and quickly adapt to novel objects.} We take Figure \ref{fm1} as an example to show the training paradigm based on multiple episode tasks. From the figure, the model needs to be trained on  $S$ and learn to infer the positions and categories of the objects from limited information in the training stage. During the test phase, the model is tasked with detecting objects from novel classes.  $Q$ consists of target classes that have not appeared in the  $S$. The model needs to leverage the learned knowledge and feature representations from the training phase to accurately detect and identify such objects.

	\begin{figure}[t]
		\vspace{0.1cm}
		\centering
		\includegraphics[width=7.32cm,height=4cm]{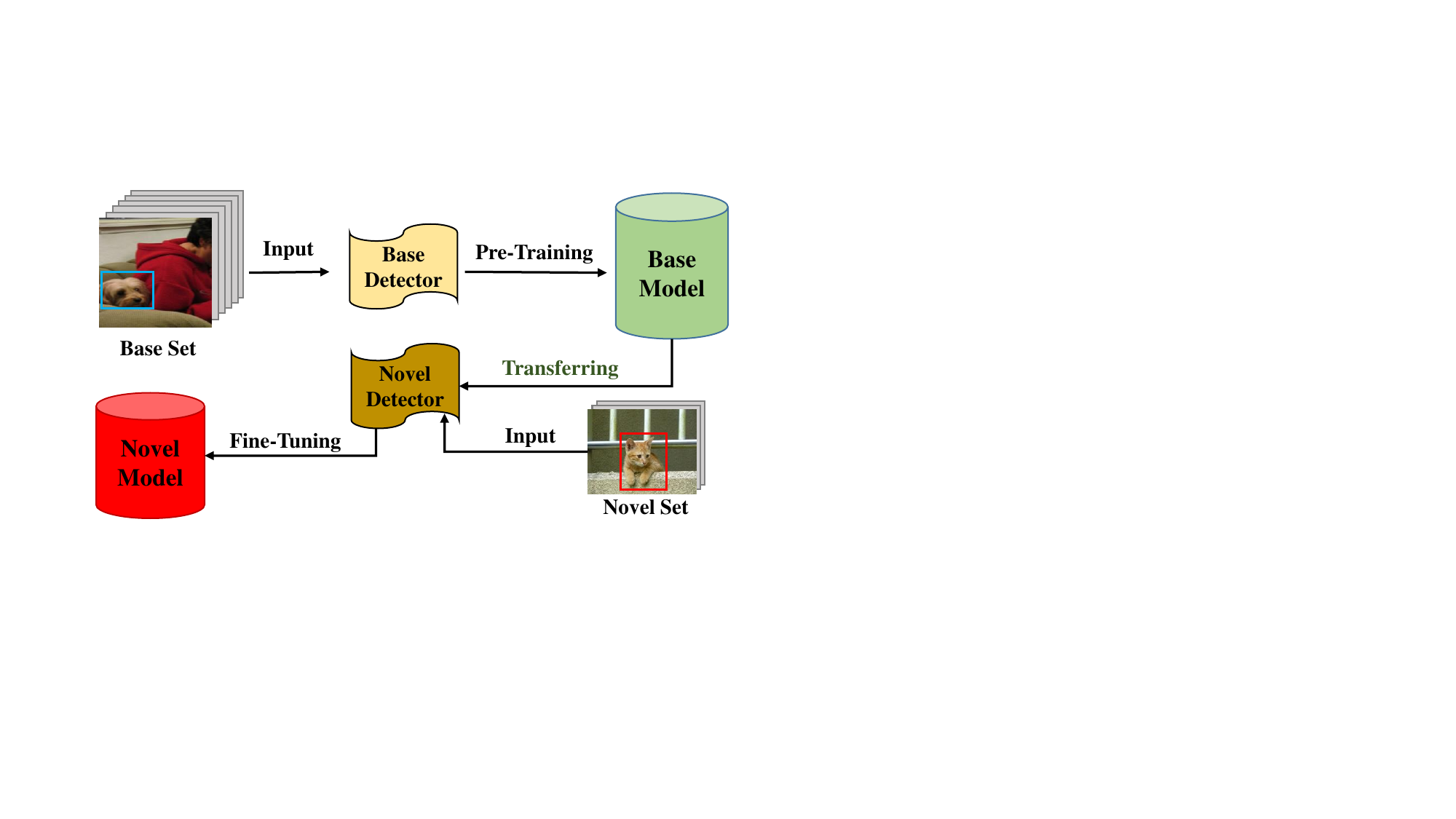} 
		\vspace{-0.3cm}
		
		\caption{
			The processing of single-task-based method.
		}

		\label{fs1}
		\vspace{-0.1cm}
	\end{figure}
	
	\subsection{Single Task based Methods}
	
	Figure \ref{fs1} illustrates the processing of single-task-based FSOD methods. From the figure, instead of multiple episode tasks, the single-task-based method is only necessary to train the traditional object detection network \cite{OB1,Fastrcnn} in the base stage without splitting the set into S and Q \cite{TFA,defrcn}. Then, the pre-trained base model is transferred to the novel detector. \textcolor{blue}{Meanwhile, the novel detector is trained on the input novel set without the need for splitting into S and Q subsets to obtain the novel model by fine-tuning \cite{lstd}, by which the model can adapt its representations and detection capabilities to the specific object categories with limited labeled examples.}

	\section{Episode Task Based Approaches}
	
	\textcolor{blue}{ In this section, we divide episode-task-based methods into four different families that use similar techniques. We then highlight the technical merits and demerits of each family. Finally, we summarize the merits and demerits of all episode-task-based methods according to their inherent characteristics.}
	
	
	\subsection{Meta-Learning for Detection Tasks}

	In \cite{metarcnn}, the authors found that meta-learning methods are very useful in few-shot classification due to only a single target needs to be identified. However, if an image contains multiple objects with complex background information, the meta-learning classification method is no longer useful because it cannot detect diverse information \cite{metalearner2}. Therefore, a series of early meta-learning-based studies \cite{metarcnn,metalearner5,metalearner,fsrw} mainly consider how to apply meta-learning to the detection task.

	\textcolor{blue}{YOLO \cite{yolov2} as a classical one-stage detection framework is adopted to combine with meta-learning in early research \cite{fsrw, metalearner3}. For example, Meta-YOLO is proposed to identify few-shot traffic signs \cite{metalearner3}. In addition, Kang \textit{et al.} \cite{fsrw} proposed FSRW, in which the YOLO head developed form \cite{yolov1} combined with meta-learning is designed to identify the object with the few-shot scenario.}
	Specifically, FSRW \cite{fsrw} is a lightweight feature reweighting module that learns to acquire the global features of support images and embeds such features into reweighting coefficients to adjust the meta features of the query image. Specifically, given a query image $ q $ and some support images $ s $ from the novel data, the feature extractor from YOLOv2 captures the meta-features of $ q $, and the reweighting module learns to obtain the global features of $ s $. Then, embed global features into the reweighting coefficient to adjust the meta-features, which assists the query meta-features in receiving support information effectively. The adaptive meta-features are entered into the detection prediction module, which is used to predict the class and bounding box of the novel object in $ q $. During the few-shot adaptation phase, FSRW focuses on reweighting the features based on the given support set, which consists of a small number of labeled examples for each object class. The reweighting process assigns higher weights to features that are more relevant for the specific few-shot detection task while down-weighting less informative or irrelevant features. Therefore, the detection performance can be improved by reweighting the features based on their relevance to each specific task or class \cite{fsrw}.

	\textcolor{blue}{However, due to the lack of accuracy and flexibility of YOLO, the framework has not been widely used in FSOD. Instead, Faster R-CNN as the two-stage detection network utilizes an RPN to generate potential object proposals, which allows for more precise localization and better handling of few-shot objects. It generally achieves higher detection accuracy compared to YOLO, which is a popular framework in the FSOD field.
	}

\begin{figure}[t]
		\vspace{0.1cm}
		\centering
		\includegraphics[width=12cm,height=6.52cm]{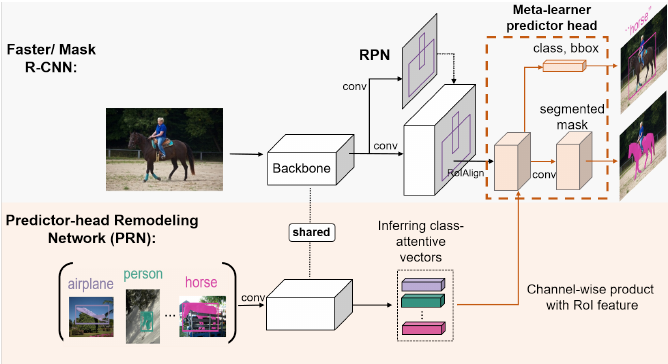} 
		\vspace{-0.3cm}
		
		\caption{
			The framework of Meta-RCNN \cite{metarcnn}.
		}

		\label{metarcnn}
		\vspace{-0.1cm}
	\end{figure}

	Figure \ref{metarcnn} as an example illustrates how to combine Faster R-CNN with meta-learning in Meta R-CNN \cite{metarcnn}. From the figure, Meta R-CNN built a meta-learner, i.e., Predictor-head Remodeling Network (PRN), that shares a common backbone of Faster R-CNN for extracting features of support images. PRN receives support images that contain few-shot objects and its bounding box to infer their class-attentive vectors. These vectors produce channel-wise with all RoI features, which reshapes the R-CNN based predictor head to detect the objects in the query image. In addition, from the figure, Meta R-CNN can also handle the few-shot segmentation task by adopting the segmented mask module \cite{maskrcnn}. In view of this, Meta R-CNN is an extension of the popular Faster R-CNN object detection framework. By using meta-learning techniques, the model becomes better at adapting its detection capabilities to new tasks and classes, especially in data scarcity scenarios. 
	
	\textcolor{blue}{ \textbf{Conclusion on meta-learning for detection tasks.} Early FSOD research primarily focused on the effective integration of the object detector with episode tasks. This fusion aims to empower the object detector to leverage the generic features acquired through FLS. Although only optimizing classification features by meta-learning limits the improvement of FSOD performance, these pioneering works laid the groundwork for subsequent developments in episode-tasks-based methods. }

	\subsection{Fusion of Episode Tasks with RPN}

	With the increasingly mature application of meta-learning in FSOD, the researchers found that some FSOD problems could be solved by improving the upstream module of the detector to enhance the final detection accuracy \cite{metaRPN1,metaRPN2}. For example, the potential bounding box can easily miss the novel object in the detection process, or produce many false detection results in the complex background \cite{metafrcn,metalearner1}. However, the above meta-learning-based FSOD works, such as FSRW \cite{fsrw} and Meta R-CNN \cite{metarcnn}, seldom consider that the anchors of the RPN output have an inappropriate low confidence score, which difficult to identify the novel objects.

	To solve this problem, Fan \textit{et al.} \cite{fan} proposed attention-RPN and a multi-relation detector. Here, attention-RPN is utilized to enhance the quality of region proposals. The multi-relation detector calculates the similarity between the support set and query set to detect novel objects while suppressing false detection in the confusing background. In addition, \cite{fan} can achieve prominent performance in FSOD without further fine-tuning.
	In \cite{fan} training strategy, as for 1-way 1-shot, the query image and support images are shared feature weights extracted by the backbone. The attention-RPN module filters out proposals that are not part of the support set and the multi-relation detector matches the query proposals with the support object. As for \textit{N-way K-shot}, the training network has $ N $ support branches, i.e., each branch needs to train its own attention-RPN and multi-relation detector. The feature of each class in the support set is the mean of $ K $ sample features.
	
	Meanwhile, Wang \textit{et al.} \cite{metadet} proposed MetaDet \cite{metadet} that simultaneously deals with few-shot classification and localization in a unified and coherent manner. Specifically, based on the Faster R-CNN framework, MetaDet \cite{metadet} adopts the meta-learning-based strategy in the detection head which is treated as category-agnostic and category-specific components. Furthermore, the category-agnostic component includes the backbone, RPN, and underlying layers of detection networks whose parameters are shared between the base and novel classes. Then, transfer such shared parameters from the base to the novel detector to implement category-agnostic meta-learning. The category-specific component contains two types of detection weights, i.e., classification and bounding box regression weights. MetaDet uses the connection of two types of weights to solve the unification problem of few-shot samples classification and localization at the same time and meanwhile extends the few-shot classification.

	In addition, Han \textit{et al.} \cite{qa} found that the RPN proposal has only a single object, which cannot be modeled with other categories. If it can be modeled, those base class features that are similar to the novel class can be utilized to enhance the feature representation of the novel class. Furthermore, due to the few-shot samples, the proposal features may have a huge difference from the class prototype features or extract uncompleted objects. To solve these problems, \cite{qa} proposed a query adaptive few-shot object detection (QA-FewDet) based on the graph convolutional networks (GCN) \cite{KG,KG2,KG3} by considering information match in the context of the image. 
	In QA-FewDet \cite{qa}, a graph is constructed to capture the relationships between the classes and proposals based on their visual features. To be specific, given a graph $ \mathbb{G} = (V, E) $, where $ V $ and $ E $ represent the node and edge of the graph, respectively. Class-specific proposals and classes become nodes in the graph, i.e., $ V_p $ for class-specific proposals of each novel class and $ V_c  $ for all base and novel classes. $ E $ is created based on the class-class, proposal-proposal, and class-proposal distances, namely $ E_{c-c} $, $ E_{p-p} $, and $ E_{c-p} $. By skillfully incorporating the GCN, QA-FewDet can enhance the discriminative power of the few-shot object detector.

	The above FSOD works abandon the negative proposals generated by the RPN module and only retain the positive proposals for detection, but such negative proposals may produce final false predictions. Therefore, appropriately restoring the negative proposal anchors generated in RPN can improve the robustness of the detectors. Based on this fact, considering the importance of the negative information in the image, the Negative- and Positive-Representative-based metric learning framework (NP-RepMet) \cite{nprepmet} was proposed to correctly recover negative information in FSOD. In the NP-RepMet \cite{nprepmet} framework, both positive and negative proposals are used in the training, in which these proposals can learn an embedded space, and metric learning is adopted to measure the similarity distance of positive and negative proposals. Specifically, negative samples are chosen to be representative of the hardest negatives, which are instances that are visually similar to the anchor sample but belong to different classes. This selection process ensures that the model learns to effectively discriminate between visually similar but class-dissimilar samples. On the other hand, positive samples are selected to be representative of the hardest positives. These are instances that are visually dissimilar to the anchor sample but belong to the same class. By including such positives, the model is encouraged to learn more fine-grained discriminative features \cite{yeshuo1,yeshuo2} that capture intra-class variations. Based on this fact, NP-RepMet \cite{nprepmet} can enhance the discriminative power of the learned embeddings by utilizing both negative and positive proposal samples during training.  The learned embedded space can be utilized in diverse FSOD frameworks to improve the performance of the data-scarcity scenario.

	The aforementioned meta learners designed for FSOD have difficulty interpreting what they are actually learning, which limits their detection performance \cite{cvpr2023}. To enhance the interpretability of meta based detectors, Demirel \textit{et al.}\cite{cvpr2023} proposed Meta-Tuning loss by adopting reinforcement learning (RL) \cite{RLear} that focuses on how a model can learn to make sequential decisions in an environment to maximize a notion of cumulative reward. Specifically, the reinforcement branch acts on the classification of the foreground and background of the RNP and the final object classification. Based on RL, the FSOD model can improve the interpretability of the object classification through trial and error to boost the generalization ability.

	\textcolor{blue}{\textbf{Conclusion on fusion of episode tasks with RPN.} Optimizing both classification and localization features by meta-learning effectively enhances detection accuracy and addresses challenges such as missed novel objects, false localization, and interpretability. However, these approaches are limited in the detection head. Due to support and query images may have diverse visual differences, e.g., background, object attitude, scale, illumination, occlusion, etc. Simply aligning such features in a high-level space may ignore the impact of the above factors \cite{fct}.}

	\subsection{Fusion of Episode Tasks with Transformer}
	
	As FSOD technology continues to mature, traditional meta-learning strategies relying solely on generic CNNs are no longer sufficient to meet the performance demands. Some researchers have turned their attention to exploring promising techniques for improving FSOD performance, such as the transformer \cite{fct,metadetr,transformerfsod10,transformerfsod9}. The transformer architecture, originally introduced for natural language processing tasks, has been adapted and extended for computer vision tasks \cite{vit,swin,detr}. This adaptation allows the transformer to effectively model spatial relationships and capture the global context in images \cite{transob1,transob2}. In addition, some fusing episode task with transformer methods \cite{transformerfsod8,transformerfsod4,transformerfsod3,transformerfsod2} have been successfully adapted for FSOD tasks. The transformer architecture brings the benefits of self-attention and global context modeling \cite{selfattention}, which can be advantageous in capturing relationships and dependencies between samples in the few-shot setting.

	CNN-based episode methods improve similarity learning by exploring the interaction between support and query images. However, the interaction of support and query features always stays in the detection head with high-level features, and the feature interaction is ignored in the feature extraction and proposal anchor stage \cite{eccv2022TENET}. Due to support and query images may have diverse visual differences, e.g., background, object attitude, scale, illumination, occlusion, etc. Simply aligning such features in a high-level space may ignore the impact of the above factors. If such features can be extracted across all network layers, the FSOD model can focus on the more common features of each layer to improve the performance of similarity learning.
	
	To achieve this goal, Han \textit{et al.} \cite{fct} proposed the fully cross-transformer (FCT) based framework, which effectively models the relationships between support and query samples by introducing cross transformer in all detection networks. Specifically, there are three cross-transformer modules in series in the backbone to extract features of the query and support set. 
	Given an input query image $I_q \in R^{1 \times H_{I_q} \times W_{I_q} \times 3} $ and $ B_S $ input support images $I_S \in R^{1 \times H_{I_S} \times W_{I_S} \times 3}$. Here, $ H $ and $ W $ represent the height and width of the image, respectively. $ B_S $ is the number of input classes and $ B_S \geq 1 $. The operation in the patch embed part is to slice the original image into non-overlapping $ 4\times4\times3 $ patches which would be mapped into $ C_1 $-dimension. The embedded patch sequence of query and support images can be given by $X_q \in R^{N_1^{q} \times C_1}$ and  $X_S \in R^{N_1^{S} \times C_1}$, where $N_1^{q}=\frac{H_{I_{q}}}{4} \times \frac{W_{I_q}}{4}$ and $N_1^{q}=\frac{H_{I_S}}{4} \times \frac{W_{I_S}}{4}$. Denote $\mathbf{E}_{q}^{pos} \in R^{N_1^{q} \times C_1}$,
	$ \mathbf{E}_{S}^{pos} \in R^{N_1^{S} \times C_1}$, and $\mathbf{E}^{bra} \in R^{2 *C_1}$
	as the position embedding of query patches, the position embedding of support patches, and branch embedding (query or support branch in meta-based FSOD), respectively. Therefore, the $ X_q $ and $ X_S $ separately combine with the position and branch encoding to obtain
	
	\begin{equation}
	X_q^{\prime}=X_q+\mathbf{E}_q^{p o s}+\mathbf{E}^{b r a}[0], 
	X_s^{\prime}=X_s+\mathbf{E}_s^{p o s}+\mathbf{E}^{b r a}[1].
	\end{equation}	
	$ X_q^\prime $ and $ X_S^\prime $ then input to the operation processing of cross-transformer that adopts PVTv2 \cite{PVTv2} to calculate the self-attention. Furthermore, FCT \cite{fct} references RoIAlign \cite{maskrcnn} to generate class-specific proposals of the query image. Finally, FCT uses the cross-transformer in stage 4 to extract RoI features for each proposal and the features of support images, respectively for final detection by pairwise matching network \cite{fan}.
	
	\textcolor{blue}{Unfortunately, the FCT network needs more computation cost than the Faster-RCNN or YOLO framework. Some researchers only adopt the transformer as the feature extractor to improve the performance of episode-based FSOD tasks. For instance, swin-transformer fused with meta-learning \cite{transformerfsod4} is proposed to detect smoke objects with data-scarcity scenarios. Zhang \textit{et al.} \cite{transformerfsod2} used the transformer after the backbone to extract high-resolution features for improving the accuracy of FSOD. Following the meta-learning concept, Li \textit{et al.} \cite{transformerfsod3} proposed a hybrid convolutional-transformer framework for drone-based FSOD.}
	
	\textcolor{blue}{In addition, some researchers only use the transformer concept in the detection head module \cite{transformerfsod8, metadetr}.} For example, Zhang \textit{et al.} \cite{metadetr} considered that Faster RCNN based FSOD may generate incomplete or even false proposals in RPN due to the data-scarcity scenario. By this, directly aggregated the features of query and support images may only predict one class in the support set, i.e., the target class to be detected. This class is essentially independent of other classes in the current support training images, which largely ignores the important inter-class dependencies between the different support classes. Thus, such FSOD detectors would have difficulty detecting similar classes, e.g., identifying cows and sheep. To solve these problems, inspired by DETR \cite{detr} and deformable DETR \cite{ddetr}, Meta-DETR \cite{metadetr} is proposed to directly utilize the inter-class relationship between the different support classes and mitigate the limitation of anchors. Deformable DETR \cite{detr} is a transformer-based model that performs end-to-end object detection by directly predicting object bounding boxes and classification from an input image, which eliminates the need for handcrafted components like RPN and anchors. Meta-DETR is an extension of the deformable DETR model \cite{ddetr} that incorporates meta-learning techniques to improve FSOD performance. Based on this fact, the DETR-based FSOD detector can skip proposal generation and directly detect novel objects at the image level. In addition, instead of repeatedly running meta-learning class by class, Meta DETR \cite{metadetr} can process the query features of multiple support classes at once. Thus, inter-class correlation between different classes can be used to reduce false classification and improve generalization. Following Mate-DETR, \cite{ transformerfsod8} utilized the encoder and decoder to improve RoI features of support and query classes. 
	
	\textcolor{blue}{\textbf{Conclusion on fusion of episode tasks with transformer.} Transformer-based methods effectively capture spatial relationships between support and query classes and global context in images, making them suitable for FSOD. However, such methods require more computation cost compared to the Faster-RCNN or YOLO framework. }

	\subsection{Exploring Interclass Relationships}
	

	The episode-task-based methods mentioned above only consider the relationship between the novel set and the detection network, ignoring the interrelation between base and novel classes in the G-FSOD setting \cite{cme}. Class appearance changes are common in FSOD \cite{vfa}. Due to the prior knowledge from the base, the novel model always learns bias toward the base class, resulting in the model easily confusing the novel object that is similar to the base class, e.g., cow as novel class is similar to sheep as base class \cite{vfa,vfa3,vfa4}. 
	
	Han \textit{et al.} \cite{vfa} consider that general FSOD approaches treat the support class as a single point in the feature space and average all features into class prototypes, which results in the true class center from few-shot samples being difficulty estimated. Therefore, inspired by the Variational Auto-Encoder (VAE) \cite{vae}, they proposed class-agnostic aggregation (CAA) and variational feature aggregation (VFA) \cite{vfa} to weaken category bias and enhance the robustness of the variances of few-shot samples, respectively, to avoid classification confusion between base and novel classes. Specifically, CAA allows feature aggregation between different categories and assists the FSOD model in learning class-agnostic representations, which reduces bias toward base classes. VFA aggregates features while incorporating the principles of variational inference, which aims to capture uncertainty or variability in the feature representations and provide more robust and expressive feature aggregation. VFA allows the FSOD model to capture the intrinsic variations and similarities between different instances within and across object classes. The support set is encoded using a deep neural network to extract discriminative features. These features capture the visual characteristics of the object classes present in the support set. The features from the support set are aggregated with the features from the query set using a variational aggregation mechanism. This aggregation process aims to capture the relevant information from the support set while considering the variations present in the query set. The aggregated features are then used for object detection in the query set. The model predicts the bounding boxes and class labels of the objects present in the query set based on the aggregated features. Based on this fact, the interaction between different categories is modeled at the same time to avoid confusion between the novel and base class.
	
	Unfortunately, \cite{vfa} ignores the relationship between the support and query classes which are treated as independent branches \cite{KFSOD}. To address this problem, Liu \textit{et al.} \cite{vfa3} proposed a dynamic graph network to fully utilize the relationship between the above classes. DCNet \cite{dcnet} is proposed to densely match support and query features. \textcolor{blue}{In \cite{metalearner6}, a dual-contrastive learning (DaCL) module is proposed to avoid object confusion in support and query sets for FSOD.} Furthermore, Lu \textit{et al.} \cite{vfa4} proposed Information-Coupled Prototype Elaboration (ICPE) method to strengthen query-perceptual information in support classes.

	On the other hand, \textbf{metric learning} is used to measure the similarity or dissimilarity between different objects, enabling the model to determine whether a novel sample belongs to a known object class or not \cite{metricfsod1,metricfsod2,zhang}. Therefore, metric learning plays a crucial role in distinguishing different categories according to their interclass relation \cite{repmet, metricfsod4,metricfsod5}. For instance,	\cite{repmet} as the earliest metric learning method proposed a DML sub-net module to improve the classification performance in detectors. In \cite{repmet}, the DML sub-net is directly added after RoI module \cite{fasterrcnn} and its output is used for classification and localization. Similarly, FM-FSOD \cite{metricfsod5} as a class-agnostic model combines meta-learning and metric learning to recognize interclass features of support and query classes. Zhang \textit{et al.} \cite{zhang} combined hybrid loss with metric learning (HLML) to score and filter proposals.	
	In addition, metric learning can also be used to detect specific objects through interclass relationships. For example, Han \textit{et al.} \cite{metricfsod1} built a metric classifier for multi-modal FSOD to identify different object features. Zhu \textit{et al.} \cite{metricfsod4} developed a meta-metric learning method to detect few-shot weld seams. Wang \textit{et al.} \cite{metricfsod5} proposed diversity measurement-based meta-learning for optical remote sensing image detection with limited samples.
	
	\textcolor{blue}{ \textbf{Conclusion on exploring interclass relationships.} Such methods consider the interrelation between different classes to learn more decision features, which effectively avoids classification confusion for FSOD when there are enough decision features. However, these methods may not work well when the object only provides the decision classification features that are similar to another object. }

	\begin{figure}[t]
		\vspace{0.1cm}
		\centering
		\includegraphics[width=12cm,height=3.23cm]{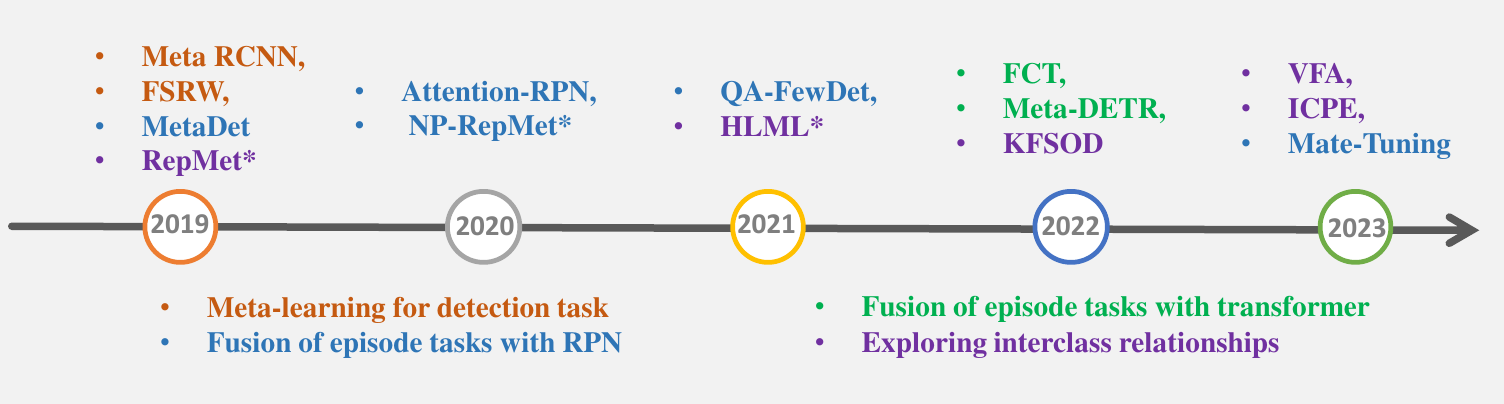} 
		\vspace{-0.3cm}
		
		\caption{
			\textcolor{blue}{Key milestones in the development of episode-task-based methods. Each color represents the same family. * represents metric learning method.}
		}

		\label{time-episode}
		\vspace{-0.1cm}
	\end{figure}
	
	\subsection{Conclusion on Episode-Task-Based FSOD Methods}
	
	\textcolor{blue}{
		\textbf{Key milestones in the development of episode-task-based methods.} We list the most iconic episode-task-based approaches that evolved within the same family in Figure \ref{time-episode}. From the figure, Meta R-CNN \cite{metarcnn} and FSRW \cite{fsrw} as the early methods that consider combining episode tasks with the classifier in the detection framework. With the increasingly mature application of episode tasks in FSOD, some methods, i.e., MetaDet \cite{metadet}, Attention-RPN \cite{fan}, QA-FewDet \cite{qa}, NP-RepMet \cite{nprepmet}, and Mate-Tuning \cite{cvpr2023}, found that by improving the upstream RPN of the detector can enhance the detection accuracy. As FSOD technology continues to mature, FCT \cite{fct} and Meta-DETR \cite{metadetr} explore transformer techniques for improving FSOD performance due to generic episode-task-based strategies relying solely on CNNs are no longer sufficient to meet the performance demands. Furthermore, exploring interclass relationships can improve the performance of FSOD. For example, RepMet \cite{repmet} and HLML \cite{zhang} utilize metric learning to measure the similarity or dissimilarity between different objects, and VFA \cite{vfa} and ICPE \cite{vfa4} aggregate support and query object features to explore interclass relationships. }
	
	\textcolor{blue}{	
		\textbf{Merit and demerit of episode-task-based FSOD methods.} 	Episode-task-based FSOD approaches follow the principle of meta-learning \cite{metal1,metal2}, in which the model can learn from limited samples and quickly adapt to novel objects by iterative training and evaluating the model on a series of episode tasks. However, such methods involve complex architectures and training procedures to increase the complexity and computational cost \cite{TFA}. In addition, they suffer from poor interpretability of what they learned in the novel stage \cite{cvpr2023}. Table \ref{meta} briefly summarizes the comparison of merit and demerit between the above FSOD methods. }

	\begin{table}[t]
		\tabcolsep=0.06cm
		\caption{Comparison of merit and demerit between episode-task-based FSOD methods. 
		}
		\scalebox{0.6}{
			\begin{tabular}{l|c|c|c}
				\midrule
				\multicolumn{1}{c|}{Methods}                    &    Pertinent Reference      & Merit                                                 & Demerit                                    \\
				
				\midrule
				
				Meta-learning for detection tasks & 
				
				
				\begin{tabular}[c]{@{}c@{}}   \cite{metalearner3}, \cite{metalearner5}, \cite{metalearner}\\  \cite{metarcnn}, \cite{fsrw},  			 \end{tabular}

				&  \begin{tabular}[c]{@{}c@{}}Rapid adaptation to\\  new tasks.  \end{tabular}                      & \begin{tabular}[c]{@{}c@{}} Limited to \\ object classification.  \end{tabular}         \\
				\midrule
				
				Fusion of episode tasks with RPN              & 
				
				\begin{tabular}[c]{@{}c@{}} \cite{metaRPN1}, \cite{metaRPN2},   \cite{fan}, \cite{metalearner1},\\ 
					\cite{metadet}, \cite{qa}, \cite{nprepmet}, \cite{cvpr2023}, \cite{metafrcn}
				\end{tabular}
				
				&  \begin{tabular}[c]{@{}c@{}} Effective distinguishing \\foreground and background.\end{tabular}   &  \begin{tabular}[c]{@{}c@{}} Limited to detection \\  head features.  \end{tabular}       \\
				\midrule

				Fusion of episode tasks with transformer  &

				\begin{tabular}[c]{@{}c@{}} \cite{fct}, \cite{detr}, \cite{transformerfsod10},  \cite{transformerfsod8},  \\ 
					\cite{transformerfsod4}, \cite{transformerfsod3}, \cite{transformerfsod2}, \cite{eccv2022TENET}	 	 
				\end{tabular}

				& \begin{tabular}[c]{@{}c@{}} Satisfying the global \\
					
					features interaction. \end{tabular}          & More computational cost.                   \\
				\midrule

				Exploring interclass relationships &
				
				\begin{tabular}[c]{@{}c@{}} \cite{vfa}, \cite{vfa3}, \cite{vfa4}, \cite{metricfsod1},  \cite{metalearner6},\\		 
					\cite{metricfsod2}, \cite{repmet}, \cite{metricfsod5}, \cite{metricfsod4}, \cite{zhang} \end{tabular}

				&\begin{tabular}[c]{@{}c@{}}  Effectively avoiding  
					
					\\ classification confusion. 
					
				\end{tabular}  & \begin{tabular}[c]{@{}c@{}}  Limited in finite  \\ decision features.  \end{tabular} \\
				\midrule
		\end{tabular}}
		\label{meta}
	\end{table}

	\section{Single Task Based Methods}
	
	Single-task-based FSOD can quickly adapt to novel classes without considering the combination of meta-learning and detector, which allows for faster convergence and enables quick deployment to significantly reduce training time in real-world scenarios. \textcolor{blue}{Furthermore, in this section, we divide single-task-based methods into five different families that use similar techniques and highlight the technical merits and demerits of each family. Finally, we summarize the merits and demerits of all single-task-based methods according to their inherent characteristics.} 	
	
	\subsection{Jointly Fine-Tuning Based Architecture}

	In jointly fine-tuning \footnote{In jointly fine-tuning training, they first train the detector on the base set and then fine-tune on the balanced dataset $ D_{balance} $ with the classes $ C_{balance} $ \cite{TFA}. }, the entire pre-trained base model, including both the class-agnostic and the class-specific layers is updated simultaneously during the training process on the novel task. All parameters of the model, both from the class-agnostic and the class-specific layers, are subject to optimization. Jointly fine-tuning allows the model to learn class-specific patterns and features while also adapting the pre-trained base knowledge to the new task. The early methods of single-task-based FSOD mainly focus on this strategy to design their detection frameworks for the few-shot scenario \cite{lstd,context}.
	
	For example, a low-shot transfer detector (LSTD) \cite{lstd} adopting this approach combines the parts from Faster R-CNN \cite{fasterrcnn} and SSD \cite{ssd} to locate and classify the objects in data-scarcity scenarios, which also is the first transfer-learning-based solution for FSOD. Specifically, LSTD replaces the bounding box regression of the original Faster R-CNN with the multi-scale box regression of SSD. The feature layer of the RoI layer \cite{fasterrcnn} is replaced by the mid-layer feature of the SSD convolution. Bounding box regression based on SSD \cite{ssd} is designed for few-shot object localization. Each convolutional layer trains box regression to generate candidate bounding boxes by using smooth L1 \cite{smoothl1}. Then, sorted the scores of the objects from the SSD regression to filter the region proposals as the result of the RPN. The middle-level convolution layer of the RoI pooling layer is used to generate a fixed-size convolution feature cube for each proposal. Instead of using the full connection layer of Faster R-CNN, two convolution layers are used in the RoI pooling layer for $ K+1 $ classification. Finally, LSTD develops the object classification based on Faster R-CNN.

	In addition, Wu \textit{et al.} proposed MPSR \cite{MPSR} to solve the problem of sparse scale distribution in FSOD based on the jointly fine-tuning strategy. MPSR as an auxiliary refinement branch generates multi-scale positive samples to further refine prediction. Furthermore, Wu \textit{et al.} \cite{MPSR} also pointed out that anchor matching on the cropped positive samples would also lead to many inappropriate negative samples, which reduces the accuracy of FSOD. To avoid this phenomenon, MPSR adopts manual selection of the corresponding scale of the feature map instead of anchor matching and takes the fixed central position as the positive sample of each object.

	As for the MPSR model optimization, denote $ N_{obj} $, $ M_{obg} $, $ L_{B c l s}^i $, and $ L_{P r e g}^i $ as the number of chosen anchors, selected positive anchor samples for refinement, the binary cross-entropy loss over objects and background, and the smooth L1 loss \cite{fasterrcnn,smoothl1}, respectively for the ith anchor in a batch image. Therefore, the model optimization of $ L_{RPN} $ in MPSR can be summarized as

	\begin{equation}
	\begin{aligned}
	& L_{R P N}=\frac{1}{N_{o b j}+M_{o b j}} \sum_{i=1}^{N_{o b j}+M_{o b j}} L_{B c l s}^i+\frac{1}{N_{o b j}} \sum_{i=1}^{N_{o b j}} L_{P r e g}^i . 
	\end{aligned}
	\end{equation}
	Then, denote $ L_{K c l s}^i $, $ M_{RoI} $, and $ N_{RoI} $ as the log loss over K classes, the number of selected RoIs, and the number of ROIs in a batch image. Therefore, the model optimization of $ L_{RoI} $ in MPSR can be given by
	\begin{equation}
	\begin{aligned}
	& L_{R o I}=\frac{1}{N_{R o I}} \sum_{i=1}^{N_{R o I}} L_{K c l s}^i+\frac{\lambda}{M_{R o I}} \sum_{i=1}^{M_{R o I}} L_{K c l s}^i+\frac{1}{N_{R o I}} \sum_{i=1}^{N_{R o I} I} L_{R r e g}^i,
	\end{aligned}
	\end{equation}
	where $ \lambda $ is a weight parameter to adjust the classification loss of the positive samples due to the fact that the number of positive samples is quite fewer than $ N_{RoI} $ in the RoI head.
	
	\textcolor{blue}{ \textbf{Conclusion on jointly fine-tuning based architecture.} Such methods first attempt to associate transfer learning with low-shot object detection, which sets a precedent for the subsequent FSOD based on the two-stage training strategy. Unfortunately, they rely particularly on situations where knowledge of the base model is highly relevant to the novel task. Therefore, jointly fine-tuning approaches are usually considered as the baseline due to their detection performance being worse than the episode-task-based methods.}
	
	\subsection{Two-Stage Fine-Tuning Approach}

	\begin{figure}[t]
		\vspace{0.1cm}
		\centering
		\includegraphics[width=12cm,height=2.86cm]{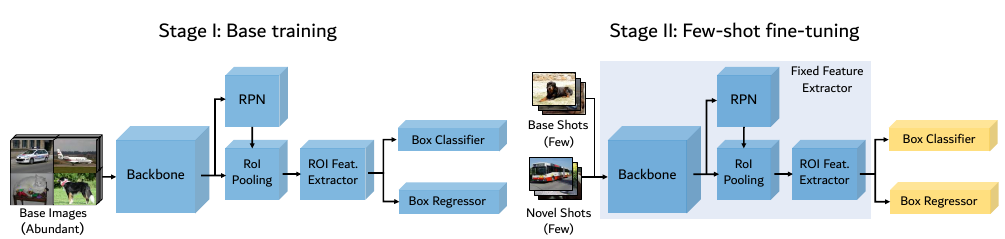} 
		\vspace{-0.3cm}
		
		\caption{
			The framework of TFA \cite{TFA}.
		} 
		
		\label{tfa}
		\vspace{-0.1cm}
	\end{figure}
	
	To overcome the problem of above methods, the two-stage fine-tuning approach (TFA) \cite{TFA} illustrates that keeping the feature extraction part of the model unchanged and only fine-tuning the last layer can improve the detection accuracy over the episode-task-based approach. This means that the feature representations learned from base classes can be transferred to novel classes.
	
	As shown in Figure \ref{tfa}, the TFA typically consists of the following two stages.
	
	\textbf{Base-training stage:} \cite{TFA} trains the entire Faster R-CNN model on the base class and adopts the same loss function. Thus, the base training loss can be given by
	
	\begin{equation}
	\mathcal{L}_{base}=\mathcal{L}_{rpn}+\mathcal{L}_{cls}+\mathcal{L}_{reg},
	\end{equation}
	where, $ \mathcal{L}_{rpn} $ is the RPN loss to distinguish the background and the proposal anchor. $ \mathcal{L}_{cls} $ and $ \mathcal{L}_{reg} $ represent the cross-entropy loss for box classification and smooth L1 \cite{fasterrcnn,smoothl1} loss for box regression, respectively. This step provides the model with a strong initial set of learned features and a general understanding of visual patterns. Then, the last layer of the base model is removed as the pre-trained model in the novel stage. 
	
	\textbf{Few-shot fine-tuning stage:} TFA \cite{TFA} randomly initializes the weights for the last layer parameters and freezes the parameters for the feature extraction module of the model, i.e., TFA only fine-tunes the bounding box classification and regression network on a class-balanced training set with few-shot samples. In training, \cite{TFA} uses a smaller learning rate and the same loss function as in the base-training stage.
	
	In addition, the updatable weight parameters in the few-shot fine-tuning stage, i.e., classification and regression network, can introduce other training strategies to robust the performance of the novel class detection. For example, TFA \cite{TFA} adds cosine similarity for box classification in this stage. Denote $ \mathcal{W} = [w_1, w_2, ..., w_c] $ as the weight matrix of the box classifier and $ \mathcal{F}(x) $ as the input further. The output of the box classifier can be scaled by similarity scores 
	
	\begin{equation}
	s_{i, j}=\frac{\alpha \mathcal{F}(x)_i^{\top} w_j}{\left\|\mathcal{F}(x)_i\right\|\left\|w_j\right\|},
	\end{equation}
	where, $ s_{i, j} $ is the cosine similarity score of $ i $th proposal feature $ \mathcal{F}(x)_i^{\top} $ and the $ j $th category weight vector $ w_j $ and $ \alpha $ represents the scaling factor. In \cite{TFA} experimental results, cosine similarity \cite{cos1,cos2} can assist FSOD model to reduce the variance between classes and improve the detection accuracy of novel classes.

	In view of this, TFA \cite{TFA} achieves competitive performance on FSOD tasks while maintaining simplicity and ease of implementation. It demonstrates that with a sample model fine-tuning, impressive few-shot detection results can be obtained. Based on this fact, some researchers might introduce variations or enhancements to the original TFA approach to further improve FSOD performance \cite{fsce,tfa2,neurips2021,eccv2022Calibration,fsodup}. For example, Sun \textit{et al.} \cite{fsce} proposed contrastive proposal encoding based on TFA, namely FSCE, to improve the performance of classification for FSOD. Specifically, the authors found that the classification task is more prone to error than localization in TFA-based FSOD works since some classes are very similar to easily cause classification confusion. They \cite{fsce} considered that contrastive learning \cite{con} is to bring similar instances closer in the learned representation space while pushing dissimilar instances further apart, which can learn useful representations from unlabeled data. Because of this, the contrastive-learning-based FSCE model \cite{fsce} is trained to discriminate between positive pairs (similar classes) and negative pairs (dissimilar classes) from the features of the RoI head by introducing contrastive learning. The loss function of this framework \cite{fsce} can be given by
	\begin{equation}
	\mathcal{L}_{fsce}=\mathcal{L}_{r p n}+\mathcal{L}_{\text {cls }}+\mathcal{L}_{\text {reg }}+\lambda \mathcal{L}_{cpe},
	\end{equation}
	where, $ \mathcal{L}_{cpe} $ is the loss of contrastive head that is proposed by \cite{fsce} and $ \lambda $ represents balance coefficient. Therefore, the model can learn to map similar pairs closer together and push dissimilar pairs apart in the learned representation space, which can enhance the feature representations of object proposals and improve the detection performance in few-shot scenarios.

	FSCE only considers the output features of RoI, but the wrong prediction of RNP leads to the failure of the features of RoI \cite{rpns}. Suppose that the detector learns to identify a category from a novel image and RPN has been trained on base classes and generates a collection of related boxes that are used to train the classifier. The classifier models have different appearances of a class by reporting multiple high IoU boxes in RPN. This means that RPN performing well in the basic category may cause serious problems for the novel class. Avoiding this effect by simply reporting a large number of boxes will result in the classifier being very good at accepting false predictions \cite{rpns}. To solve the problem, Sun \textit{et al.} \cite{rpns} proposed a strategy for learning cooperating RPN’s (CoRPNs). While this may seem redundant when one RPN fails to predict a high IoU, another RNP can successfully predict it. This strategy also avoids overpredicting to some extent.

	The performance of the generated proposal also influences TFA. For example, Ma \textit{et al.} \cite{eccv2022} found that the deep RoI features may predict too many redundant background features, which reduces training efficiency and affects the detection accuracy. To reduce the appearance of redundant features in the proposal, they proposed CoCo-RCNN which can assist the model to concentrate foreground features and degrade the background. In addition, Fan \textit{et al.} \cite{eccv2022Calibration} found except for many redundant features in the proposal, and existing incomplete objects in the proposal. The presence of abundant such prediction proposals will confuse the RPN classifier and lead to false positives. Kual \textit{et al.} \cite{cvpr2022} proposed a verification technique to remove the above controversial proposal prediction in RPN by self-supervision training and enhance the quality of the bounding box by training a specialized regressor. Unfortunately, RPN difficulty generates proposal boxes when the number of shots in a class is 3 or less, which results in objects being unpredictable. To solve the problem, Zhang \textit{et al.} \cite{hall} proposed a hallucinator network to generate additional and useful training samples in the feature space of RoI.
	

	\textcolor{blue}{ \textbf{Conclusion on two-stage fine-tuning approach.}	TFA-based methods are known for their faster convergence and quick deployment to significantly reduce training time in novel scenarios. However, differences in data distribution between the base domain and the novel domain may lead to a bias in the training model using the base domain on the novel domain, which degrades the performance of the few-shot model. To solve this problem, most of the subsequent single-task-based methods combined with knowledge distillation, context reasoning, or decoupling to improve the detection performance, but still follows the learning paradigm of TFA. For a clearer overview, we will bring together the methods using the same techniques in a subsection.}
	
	\subsection{Knowledge Distillation}


	A potential assumption of the above TFA-based paradigm is that the prior knowledge unrelated to the class can be implicitly transferred from the base to novel classes, which makes it difficult for the FSOD model to use the common features between the base and novel classes. To solve the problem, Sun \textit{et al.} \cite{eccv2022multifacedistill} proposed a scheme that can learn the multifaceted commonalities between base and novel classes by distilling the commonality knowledge \cite{kl,niff}. 
	
	Specifically, knowledge distillation \cite{kl} is used to transfer knowledge from a complex teacher model to a simpler student model. It improves the performance of the student model by leveraging the knowledge acquired by the teacher model during training. In \cite{eccv2022multifacedistill}, the teacher (base) model is typically trained on a larger dataset and the student (novel) model is trained on few-shot samples to mimic the behavior of the teacher model while being computationally efficient. By distilling knowledge from the base to the novel model, the FSOD model can benefit from the base's expertise, generalization capabilities, and learned representations. This allows the FSOD model to achieve better performance in novel classes with limited annotated data, making it more effective in scenarios where acquiring large amounts of labeled data is challenging or impractical. Therefore, \cite{eccv2022multifacedistill} based on the TFA framework explicitly learns the multifaceted commonalities between the base and novel classes.	Meanwhile, Pei \textit{et al.} \cite{eccv2022kdisll} also found potential class-specific overfitting on base and novel classes due to ignoring the common features between them. They perform knowledge distillation on the bag of visual words which is used to learn the similarities of objects to teach the learning of object detectors.
	
	However, \cite{eccv2022kdisll} and \cite{eccv2022multifacedistill} distill the knowledge learned from the teacher (base) model directly into the student (novel) classes, which may transfer the wrong prediction from the teacher to the student model. To solve the problem, Li \textit{et al.} \cite{AAAI2023Kdistillation2} proposed a structural causal model between the base and novel model to perform a conditional causal intervention, which can prevent error prediction from the teacher stage to ensure detection accuracy.

	Furthermore, Jiang \textit{et al.} \cite{knowdistillfsod1} found that the performance of the FSOD framework could be improved by knowledge distillation. They introduced the mutual distillation layer into the Faster R-CNN detection head to alleviate the generation of redundant anchors. Based on this fact, Li \textit{et al.} \cite{knowdistillfsod2} introduced the self-knowledge distillation algorithm into the Faster R-CNN framework to improve the performance of foreground detection. Nguyen \textit{et al.} \cite{eccv2022Counting} adopted knowledge distillation to fine-tune the DETR framework.
	
	
	\textcolor{blue}{ \textbf{Conclusion on knowledge distillation.}
		During the process of knowledge distillation, the teacher (base) model's output can serve as an additional supervised signal to guide the training of the student (novel) model. This invaluable base information aids the novel model in better understanding and capturing the key features to improve the performance of FSOD. However, it relies on abundant similarity features between base and novel classes, otherwise, it is easy to lead to wrong teaching \cite{knowdistillfsod1}. In addition, learning multifaceted commonalities between base and novel classes will add extra computation to cost more training resources \cite{eccv2022multifacedistill}.
	}


	\subsection{Context Reasoning}
	
	\textcolor{blue}{To mitigate the over-reliance on similar features between the base and novel classes, context-reasoning-based FSOD methods solely focus on utilizing the features within an image to deduce target categories based on context \cite{context,SRR}, thereby enhancing the robustness of the novel model.} Specifically, the automatic exploitation of contextual information from the few images and careful integration of these different clues can assist the model in clear detection, i.e., attempts to explore different clues in the surrounding environment to clarify category confusion. For example, some images may sufficiently distinguish a horse from a dog when a person sitting on the animal and the scene is about a racetrack. 
	
	Based on the above motivation, Aditya \textit{et al.} \cite{context} proposed a context transformer to alleviate the categories confusion in the classification task due to the low data diversity. Context transformer is composed of affinity discovery and context aggregation as the plug-and-play submodules embed into the SDD-style detector \cite{ssd}. Given an input image, affinity discovery first builds a set of context fields based on the default anchor box in the SDD detector and adaptively exploits the relationship between the anchor box and the context field. Then, context aggregation takes advantage of these relationships to centralize the key context into each anchor box. Therefore, the context converter can generate a context-aware representation for each anchor box, which allows the detector to distinguish between few-shot obfuscation by discriminating context clues.

	
	On the other hand, the application of knowledge graphs (KG) \cite{KG, KGraph22,KGraph33} in the context involves leveraging structured knowledge to improve the performance of object detection models when dealing with limited training data. Specifically, KG \cite{SRR} provides information about the semantic relationships between object categories. This information guides the model to understand the associations between different object classes, which can be particularly useful when dealing with few-shot scenarios. For example, if the KG indicates that "car" and "road" are often related, this knowledge can help the model when detecting cars on roads.

	As for the KG application, \cite{SRR} and \cite{contextfsod5} incorporate semantic relation reasoning in the image to improve the performance of the model. They leverage the semantic relations and contextual information between objects to better understand their connections and dependencies. Zhu \textit{et al.} \cite{SRR} proposed SRR-FSD that utilizes graph modeling techniques \cite{KG}, attention mechanisms, and relational reasoning to model the semantic relations between objects. SRR-FSD can construct a graph of relationships, where nodes represent target instances and edges represent their associations. The model can then perform inference and information propagation on the graph to obtain comprehensive contextual information and achieve more accurate object detection in few-shot scenarios.
	
	With regard to the processing of relation reasoning \cite{SRR}, in the second stage of Faster R-CNN, the feature vector of the RoI is extracted and transformed into the D-dimensional vector $ \mathbf{v}\in R^d $ by the fully connected layer in the classification subnet. Denote $ \mathbf{W}\in R^{\mathbf{N}\times d} $ and $ \mathbf{b}\in R^\mathbf{N} $ as learnable weight matrix and learnable bias vector, respectively, where $ \mathbf{N} $ represents the number of classes. Then, the probability distribution \cite{SRR} in the original classification task can be given by	
	\begin{equation}
	\mathbf{p}=\operatorname{softmax}\left( \mathbf{W} \mathbf{v}+\mathbf{b}\right).
	\label{1}
	\end{equation}	
	To learn semantic association, the visual feature $ \mathbf{v} $ is projected into the semantic space, which adopts $ d_e $-dimensional word embeddings, referred to as $ \mathbf{W}_e\in R^{\mathbf{N} \times d_e} $. The classification subnet of the detector learns a linear projection matrix $ \mathbf{P}\in R^{d_e\times d} $. By aligning the visual feature $ \mathbf{v} $ with the word embeddings, the probability \cite{SRR} of formula (\ref{1}) becomes
	\begin{equation}
	\mathbf{p}=\operatorname{softmax}\left( \mathbf{W}_e \mathbf{P} \mathbf{v}+\mathbf{b}\right).
	\label{2}
	\end{equation}	
	When learning a new category, only $ \mathbf{W}_e $ needs to be extended and the projection matrix $ \mathbf{P} $ remains unchanged. Relational reasoning transfers knowledge between classes. The relation reasoning graph $ \mathbf{G} $ is the $ N \times N $ adjacency matrix, which represents the connection strength between each class and the other classes. Thus, formula (\ref{2}) in \cite{SRR} is updated as
	\begin{equation}
	\mathbf{p}=\operatorname{softmax}\left(\mathbf{G} \mathbf{W}_e \mathbf{P} \mathbf{v}+\mathbf{b}\right).
	\end{equation}	
	By incorporating semantic relation reasoning, the SRR-FSD method \cite{SRR} enables the utilization of relational information between objects to compensate for the limited number of samples, thereby improving the robustness and generalization ability of the model. This approach allows the model to handle FSOD tasks more effectively and adapt well to novel and unseen classes.

	In addition, Aditya \textit{et al.} \cite{SR} considers that the co-occurrence of objects in spatial information with base categories, the feature representation of novel categories can be inferred. By this, they found that the interrelationship between objects is particularly important and useful for the FSOD model to explicitly learn the location of several objects in an image. \cite{SR} proposed a few-shot object detector with a spatial reasoning framework (FSOD-SR) that can predict the novel object in a context. FSOD-SR defines a spatial graph after the Faster RCNN RoI stage. FSOD-SR uses a spatial graph to enhance the discriminant ability of foreground information in an image. In this graph, nodes are the training categories in the base or novel stage and edges represent the relatedness of each node.
	
	On the other side, the FSOD with context reasoning technique has been widely used in remote sensing \cite{contextfsod1,contextfsod2} and the combination of text and image fields \cite{contextfsod4,contextfsod3}. For example, Wang \cite{contextfsod1} \textit{et al.} found that the generic FSOD detector makes it difficult to identify objects in remote sensing images with diverse backgrounds and different object sizes. To solve the issue, They proposed a CIR-FSD module based on the context information of such images to capture the related object features from different receptive fields. This module can assist RPN in relaxing the constraint of non-maximum suppression (NMS) \cite{NMS} in the fine-tuning stage. Furthermore, Yu \cite{contextfsod3} \textit{et al.} proposed a knowledge-augmented technique to boost the generalization ability of the model for few-shot visual relation detection.

	\textcolor{blue}{ \textbf{Conclusion on context reasoning.}
		By focusing on the internal features of the image, the context-based FSOD methods can reduce the impact of similarity between categories, which enables the model to distinguish different object categories accurately. Unfortunately, such FSOD methods apply to objects with multiple relationships in an image. It is difficult to build relation reasoning with semantic relationships on the graph or context when there is only one object in an image with a popular background.}

	\subsection{Decoupling Detection Networks}
	
	Decoupling the detection network can enhance the generalization ability of the FSOD model independently of the semantic relationship between classes and backgrounds. This technology refers to separating or reducing the interdependencies between different components or aspects of the model, such as its architecture, parameters, or input/output mechanisms to increase modularity and flexibility. It allows for easier modifications, enhancements, or adaptations to different domains, especially in FSOD tasks \cite{defrcn}.
	
	In \cite{defrcn}, the authors found the following three problems in FSOD. Since most parameters are pre-trained on the base set and then frozen on the novel set, the FSOD model can suffer from a severe shift in data distribution and low utilization of novel data. Secondly, the class-agnostic RPN and class-specific RCNN are jointly optimized through a shared backbone, which leads to conflicts between the three modules. In addition, the features in the box regressor and box classifier of RCNN are different, which severely limits the performance of the object classification.
	
	\begin{figure}[t]
		\vspace{0.1cm}
		\centering
		\includegraphics[width=12cm,height=2.38cm]{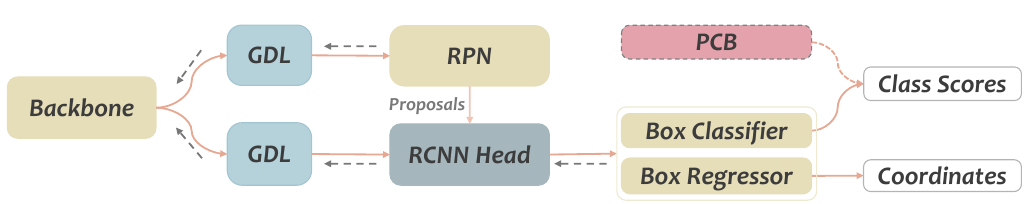} 
		\vspace{-0.3cm}
		
		\caption{
			The framework of DeFRCN \cite{defrcn}.
		} 
		
		\label{defrcn}
		\vspace{-0.1cm}
	\end{figure}
	
	To solve these three problems, \cite{defrcn} proposed decoupled Faster R-CNN (DeFRCN) to cover the shortage of Faster R-CNN based detectors in FSOD. Specifically, DeFRCN involves a Gradient Decoupled Layers (GDL) module which adds between backbone, RPN, and RCNN Head, as shown in Figure \ref{defrcn}. In the forward and backward propagation process, GDL performs the affine transformation layer of the forward feature. In the backward, GDL employs the subsequent gradient to multiply a constant. Based on this fact, the DeFRCN model achieves decoupling between the multiple stages. Denote $ F(\cdot) $ and $ \mathbb{G} $ as the function of the FSOD detector and GDL layer, respectively. The optimization process based on the decoupling of the above three modules can be given by the following loss
	\begin{equation}
	\begin{aligned}
	\mathcal{L}_{defrcn}= & \mathcal{L}_{r p n}\left(F_{r p n}\left(\mathbb{G}_{\boldsymbol{r p n}}\left(F_b\left(x ; \theta_b\right)\right) ; \theta_{r p n}\right), y_{r p n}\right)+\eta \cdot \\
	& \mathcal{L}_{r c n n}\left(F_{r c n n}\left(\mathbb{G}_{\boldsymbol{r c n n}}\left(F_b\left(x ; \theta_b\right)\right) ; \theta_{r c n n}\right), y_{r c n n}\right)
	\end{aligned}
	\end{equation}
	where, $ \theta_b $, $ \theta_{RPN} $ and $ \theta_{RCNN} $ represent the learnable parameters of backbone, RPN, and RCNN, respectively. Furthermore, $ \eta $ is a hyper-parameter to balance $ \mathcal{L}_{rpn} $ and $ \mathcal{L}_{rcnn} $. In addition, DeFRCN adopts the prototypical calibration block (PCB) \cite{defrcn} block to improve the performance of the object classification. 
	
	This work achieved milestone performance in FSOD, and several studies extended their work follow-up on the decoupled concept. For example, \cite{tip1} \textit{et al.} build PTF+KI upon DeFRCN without the PCB module to improve the efficiency and accuracy of FSOD. Yan \textit{et al.} \cite{defrcn2} exploited a model decoupling technique to prevent the phenomenon of proposal estimation bias in FSOD. In addition, decoupled detection network methods have been utilized in detecting specific objects. For instance, Zhang \textit{et al.} \cite{defrcn3} decoupled the remote sensing image detector to keep the performance of the base model and improve the generation of the novel model. Su \textit{et al.} \cite{defrcn1} decouples the novel detector for few-shot open-set object detection to avoid overfitting.

	
	On the other hand, adopting the two-stage detection framework by using RoI to predict the object may only recognize a single object resulting in inefficiency. To solve the limitation, Lu \textit{et al.} \cite{dmnet} proposed a decoupling paradigm to achieve image-level processing and high efficiency based on the one-stage detector. \cite{dmnet} includes a decoupling representation transform (DRT) module. Specifically, DRT first distinguishes the foreground and background to extract foreground representation. DRT then predicts the shape of the anchor to solve the problem that due to the handcrafted anchor being far from the ground truths, the difficulty of regression and the inadequacy of training positive samples would be increased in one-stage detectors. Furthermore, DRT can obtain task-related features by adapting features for classification and location. In view of this, several adaption modes in DRT are designed to decouple the representation of classification and location features to obtain both appropriate receptive fields. Finally, the DRT is designed on the pipeline of one-stage detectors \cite{ssd,yolov1,yolov2} to further boost the generalization ability.

	\textcolor{blue}{ \textbf{Conclusion on decoupling detection networks.}
		By decoupling the detection modules, the model can effectively capture class-specific information for both base and novel classes, thereby significantly enhancing its generalization capabilities. While these methods enable the model to achieve impressive performance in FSOD, it is important to note that they necessitate the incorporation of additional mechanisms to facilitate the decoupling process. This, in turn, leads to increased architectural complexity and training procedures, potentially resulting in heightened computational requirements and longer training times.}


	\begin{table}[t]
		\tabcolsep=0.06cm
		\caption{Comparison of merit and demerit between single-task-based FSOD methods. 
		}
		\scalebox{0.55}{
			\begin{tabular}{l|c|c|c}
				\midrule
				\multicolumn{1}{c|}{Methods}             & Pertinent Reference &  Merit                                             & Demerit                                     \\
				\midrule

				Jointly fine-tuning based architecture		
				& \cite{lstd}, \cite{MPSR}, \cite{context} &      
				\begin{tabular}[c]{@{}c@{}}Rapid adaptation to\\ new tasks. \end{tabular}

				& 
				\begin{tabular}[c]{@{}c@{}}
					Low precision and rely on highly \\ relevant between base and novel classes. \end{tabular}                             \\
				\midrule
				
				Two-stage fine-tuning approaches
				
				&  \begin{tabular}[c]{@{}c@{}}
					\cite{TFA}, \cite{fsce}, \cite{tfa2}, \cite{neurips2021}, \cite{eccv2022Calibration}, 
					\\
					\cite{fsodup}, \cite{rpns}, \cite{eccv2022}, \cite{cvpr2022}, \cite{hall}
				\end{tabular}

				&  \begin{tabular}[c]{@{}c@{}}Faster training and \\ easy deployment.   \end{tabular}              & \begin{tabular}[c]{@{}c@{}} Severe shift in domain target. \end{tabular}             \\
				
				\midrule

				Knowledge distillation        &  
				
				\begin{tabular}[c]{@{}c@{}}
					\cite{eccv2022multifacedistill}, \cite{niff}, \cite{eccv2022kdisll}, \cite{AAAI2023Kdistillation2},
					\\
					\cite{knowdistillfsod1}, \cite{knowdistillfsod2}, \cite{eccv2022Counting}
					
				\end{tabular} 
				&  
				\begin{tabular}[c]{@{}c@{}}Providing additional  \\  supervised signals. \end{tabular} 
				
				& \begin{tabular}[c]{@{}c@{}} Rely on abundant similarity \\  features  between $ C_{base} $ and $ C_{novel} $.  \end{tabular}         \\
				\midrule

				Context reasoning         &  
				
				\begin{tabular}[c]{@{}c@{}}
					\cite{context}, \cite{SRR}, \cite{contextfsod5}, \cite{SR},
					\\
					\cite{contextfsod1}, \cite{contextfsod2}, \cite{contextfsod4}, \cite{contextfsod3}  
					
				\end{tabular} 
				&  
				\begin{tabular}[c]{@{}c@{}}Suitable for complex\\ relationship modeling. \end{tabular} 
				
				& \begin{tabular}[c]{@{}c@{}} Limited in the images with 
					\\
					relationship features.  \end{tabular}         \\
				\midrule

				Decoupling detection networks      &  
				
				\begin{tabular}[c]{@{}c@{}}
					\cite{defrcn}, \cite{tip1}, \cite{cfa}, \cite{niff},
					
					\cite{ecea}, \\ \cite{dmnet}, \cite{defrcn3}, \cite{defrcn1}, \cite{defrcn2}  
					
				\end{tabular}  
				
				& \begin{tabular}[c]{@{}c@{}} Strong generalization \\ ability. \end{tabular}                   & 	\begin{tabular}[c]{@{}c@{}} Increased model complexity 
					
					
				\end{tabular} \\

				\midrule
				
				Keeping the performance on base categories &

				\begin{tabular}[c]{@{}c@{}}
					\cite{cfa}, \cite{cvpr2023gfsod}, \cite{retina}, \cite{niff},
					
					\cite{increfsod1}, \\ \cite{increfsod2}, \cite{increfsod3}, \cite{increfsod4}, \cite{increfsod5}  
					
				\end{tabular}

				&  
				\begin{tabular}[c]{@{}c@{}}Avoiding catastrophic \\ forgetting of base classes. \end{tabular}
				& \begin{tabular}[c]{@{}c@{}} complexity of \\ implementing deployments.  \end{tabular}                 \\

				\midrule
		\end{tabular}}
		\label{single}
	\end{table}

	\subsection{Keeping the Performance on Base Categories}

	
	The above model trained on a novel set of classes may lose its ability or perform poorly to recognize and detect objects from the base classes. To overcome this challenge, researchers have proposed G-FSOD techniques to mitigate catastrophic forgetting in few-shot detection. G-FSOD without forgetting is an extension of the FSOD task that aims to train a model that can not only detect objects in novel classes with limited labeled data but also retain its ability to detect objects in previously seen classes without significant performance degradation \cite{cfa, cvpr2023gfsod}.
	
	Motivated by this, Fan \textit{et al.} \cite{retina} proposed Retentive R-CNN to eliminate the catastrophic forgetting of base classes. Retentive R-CNN includes Bias-Balanced RPN and Re-detector. To be specific, Bias-Balanced is used to ensure that the effects of the base class are not catastrophic forgetting by integrating base pre-trained RPN and novel RPN. Denote $ O_b $ and $ O_n $ as the objects of base RPN prediction and finetuned RPN, respectively. The output of the final Bias-Balanced RPN in \cite{retina} is defined as $ O=(O_b, O_n) $. Furthermore, in the fine-tuning stage, only the finetuned RPN module is unfrozen. On the other hand, the Re-detector includes a base RoI detector in the RoI module to predict base classes and the novel RoI detector to detect base and novel classes. \cite{retina} adopts consistency loss in this module, which can be applied to encourage consistency between different views of the same input detector. The idea of consistency loss enforces that the model's predictions remain stable or consistent regardless of the specific perturbation applied to the input. In addition, in the fine-tuning stage, the novel RoI detector is unfrozen.
	
	
	
	If the fine-tuning stage only provides novel classes without any base class to train the detection network, how it avoids catastrophic forgetting? For this problem, Guirguis \textit{et al.} \cite{niff} proposed Neural Instance Feature Forging (NIFF) to alleviate catastrophic forgetting without storing base data. NIFF trains an independent lightweight generator by knowledge distillation to synthesize class-wise base features for G-FSOD training.

	
	On the other side, incremental learning can also address catastrophic forgetting by allowing models to learn incrementally, incorporating new data without discarding previously acquired knowledge. FSOD combined with this technique can mitigate or minimize catastrophic forgetting of base classes and improve the performance of novel classes. There are a series of related incremental learning methods \cite{increfsod1, increfsod2, increfsod3, increfsod4, increfsod5} to ensure the model retains knowledge from base learning. For example,  Feng \textit{et al.} \cite{increfsod1} proposed MCH and BPMCH models based on incremental learning to create connections between base and novel classes, which maintain the base performance and transfer more knowledge from base to novel class. Choi \textit{et al.} \cite{increfsod3} proposed an incremental two-stage fine-tuning approach, in which the incremental module is designed for RoI features to preserve base regression and classification tasks. In addition, Deng \textit{et al.} \cite{increfsod5} proposed a class-incremental FSOD framework to continually learn new samples without forgetting previous training in practical robotics applications.

	\textcolor{blue}{ \textbf{Conclusion on keeping the performance on base categories.}
		G-FSOD methods are designed to prevent catastrophic forgetting of the performance of base objects during the learning process of a novel model. Such methods offer the advantage of handling multiple categories in a flexible and scalable manner, allowing models to adapt to various practical application scenarios. However, to maintain knowledge of known base objects, additional steps are required to manage and update the model's weights. This can lead to increased computational requirements and complexity when implementing these methods in real-world deployments.}

	
\begin{figure}[t]
		\vspace{0.1cm}
		\centering
		\includegraphics[width=12cm,height=3.77cm]{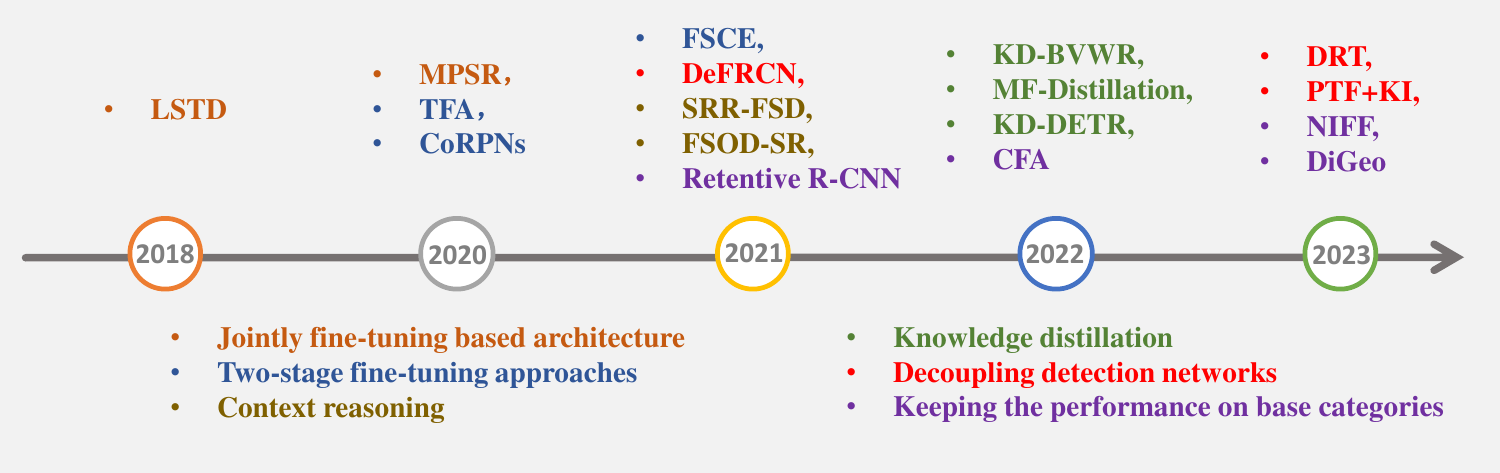} 
		\vspace{-0.3cm}
		
		\caption{
		{\color{blue} Key milestones in the development of single-task-based methods. Each color represents the same family.}
		} 		
		\label{time-single}
		\vspace{-0.1cm}
	\end{figure}
	
	\subsection{Conclusion on Single-Task-Based FSOD Methods}
	
	\textcolor{blue}{
		\textbf{Key milestones in the development of single-task-based methods.} Figure \ref{time-single} illustrates the most iconic single-task-based approaches that evolved within the same family. LSTD \cite{lstd} is the earliest method to propose the two-stage training strategy for FSOD and with its extending method, MPSR \cite{MPSR} follows the jointly fine-tuning concept. However, such methods are usually regarded as the baseline with low performance. Therefore, in 2019, most approaches began to consider how to integrate them with episode tasks. Until 2020, TFA \cite{TFA} broke the stagnation of single-task development with its fast and accurate training strategy. By this, TFA as the new baseline of single task method extends FSCE \cite{fsce}, CoRPNs \cite{rpns}, etc. Besides, most of the subsequent TFA-based methods combined with knowledge distillation, e.g., KD-BVWR \cite{eccv2022kdisll}, MF-Distillation \cite{eccv2022multifacedistill}, and KD-DETR \cite{eccv2022Counting}
		context reasoning, e.g., SRR-FSD \cite{SRR} and FSOD-SR \cite{SR}, or decoupling detection networks, DeFRCN \cite{defrcn}, DRT \cite{dmnet}, and PTK+KI \cite{tip1}. Furthermore, Retentive R-CNN \cite{retina} with its extending methods, CFA \cite{cfa}, NIFF \cite{niff}, and DiGeo \cite{cvpr2023gfsod} considers that keeping the base performance in the G-FSOD protocol can provide more comprehensive results reporting.}
	
	\textcolor{blue}{
		\textbf{Merit and demerit of single-task-based FSOD methods.} These single-task-based FSOD approaches with faster training and simple deployment can quickly adapt to novel classes and achieve strong generalization ability with the ingenious decoupling technique. Unfortunately, there is no uniform fine-tuning strategy, in which different FSOD tasks freeze or unfreeze different detection modules \cite{cfa}. In addition, if the novel domain significantly deviates from the base domains used during training, the performance may degrade due to the domain gap \cite{defrcn}. Table \ref{single} briefly summarizes the comparison of merit and demerit between the above FSOD methods.}

	\begin{table}[t]
		\tabcolsep=0.08cm
		\caption{Comparison of merit and demerit between episode task and single task FSOD methods. 
		}
		\scalebox{0.7}{
			\begin{tabular}{l|c|c}
				\midrule
				\multicolumn{1}{c|}{taxonomy}             & Merit                                                                                                    & Demerit                                                                                                                     \\
				\midrule
				Episode Task Methods & \begin{tabular}[c]{@{}c@{}}Strong adaptive ability without fine-tuning.\\ Mitigating domain shift.\end{tabular}        & \begin{tabular}[c]{@{}c@{}} Poor interpretability of what is learned. \\ Complexity and more computational cost.\end{tabular} \\
				
				\midrule
				Single Task Methods & \begin{tabular}[c]{@{}c@{}}Faster training and deployment.\\ Strong generalization ability.\end{tabular} & \begin{tabular}[c]{@{}c@{}}Severe shift in domain target.\\ Difficult designing of fine-tuning details.\end{tabular}  \\
				\midrule     
		\end{tabular}}
		\label{mateandsingle}
	\end{table}

	\begin{table*}[t]
		\centering	
		\tabcolsep=0.09cm
		
		\caption{Performance of FSOD on three PASCAL VOC novel sets (AP50). Symbol `-' represents unreported results in the original work. Bold font indicates the best result in the group. FRCN (R101) and FRCN (R50) are Faster R-CNN with the backbone of ResNet-101 or ResNet-50. DK19 represents Darknet19 from \cite{yolov2} and VGG16 is the visual geometry group network \cite{vgg16}. D-DETR is deformable DETR \cite{ddetr}.
		}
		\scalebox{0.6}{
			\begin{tabular}{@{}ll|c|ccccc|ccccc|ccccc@{}}
				\toprule
				& \multirow{2}{*}{Methods} 	& \multirow{2}{*}{Framework} &   \multicolumn{5}{c|}{Novel Split1}                                                                                    & \multicolumn{5}{c|}{Novel Split2}                                                                                    & \multicolumn{5}{c}{Novel Split3}                                                                                                         \\ 
				& &	&                   \multicolumn{1}{c}{K=1}                         & \multicolumn{1}{c}{2}                         & \multicolumn{1}{c}{3}                         & \multicolumn{1}{c}{5}    & \multicolumn{1}{c|}{10}    & \multicolumn{1}{c}{1}                         & \multicolumn{1}{c}{2}                         & \multicolumn{1}{c}{3}                         & \multicolumn{1}{c}{5}    & \multicolumn{1}{c|}{10}     & \multicolumn{1}{c}{1}                         & \multicolumn{1}{c}{2}                         & \multicolumn{1}{c}{3}                         & \multicolumn{1}{c}{5}    & \multicolumn{1}{c}{10}                         \\ \midrule

				& \textit{Episode Task}   \\
				
				&	\multicolumn{1}{l|}{MetaDet~\cite{metadet}}   & FRCN (VGG16)                       &  18.9                      & 20.6                      & 30.2                      & 36.8 & 49.6 & 21.8                      & 23.1                      & 27.8                      & 31.7  & 43.0 & 20.6                      & 23.9                      & 29.4                      & 43.9                      & 44.1                      \\ 
				
				&  FSRW \cite{fsrw} &  YOLOv2 (DK19) & 14.8  & 15.5  & 26.7  & 33.9  & 47.2  &  15.7  & 15.3  & 22.7  & 30.1  & 40.5  & 21.3  & 25.6  & 28.4  & 42.8 &  45.9 \\
				
				&	\multicolumn{1}{l|}{Meta R-CNN~\cite{metarcnn}}      & FRCN (R101)                &  19.9                      & 25.5                      & 35.0                      & 45.7 & 51.5 & 10.4                      & 19.4                      & 29.6                      & 34.8 & 45.4 & 14.3                      & 18.2                      & 27.5                      & 41.2                      & 48.1                      \\ 
				
				&  FSDetView \cite{xiao}  &   FRCN (R101)   &  24.2   &  35.3   &  42.2   &  49.1   &  57.4   &  21.6  &   24.6  &   31.9   &  37.0   &  45.7   &  21.2   &  30.0   &  37.2  &   43.8  &   49.6 \\
				
				&	QA-FewDet~\cite{qa}   &FRCN (R101) & 42.4 &51.9 &55.7 &62.6 &63.4 &25.9 &37.8 &46.6 &48.9 &51.1 &35.2 &42.9 &47.8 &54.8 &53.5\\
				
				&	Meta FRCN~\cite{metafrcn}  & FRCN (R101) & 43.0 & 54.5 &60.6 &66.1 &65.4 &27.7 &35.5 &46.1 &47.8 &51.4 &40.6 &46.4 &53.4 &59.9 &58.6 \\
				
				&  NP-RepMet \cite{nprepmet} &  FRCN (R101)  &   37.8  &   40.3  &   41.7   &  47.3   &  49.4   &  41.6   &  43.0   &  43.4   &  47.4   &  49.1   &  33.3   &  38.0  &   39.8  &   41.5   &  44.8 \\

				&	FCT~\cite{fct}   & FRCN (PVTv2) &49.9 &57.1 &57.9 &63.2 &67.1 &27.6 &34.5 &43.7 &49.2 &51.2 &39.5 &54.7 &52.3 &57.0 &58.7\\
				
				& DCNet \cite{dcnet} & FRCN (R101) &33.9  & 37.4 &43.7 &51.1 &59.6 &23.2 &24.8 &30.6 &36.7 &46.6 &32.3 &34.9 &39.7 &42.6 &50.7 \\
				
				&CME \cite{cme} & YOLOv2 (DK19) & 17.8 &26.1 &31.5 &44.8 &47.5 &12.7 &17.4 &27.1 &33.7 &40.0 &15.7 &27.4 &30.7 &44.9 &48.8 \\
				
				&CME \cite{cme} & FRCN (R101) &41.5 &47.5 &50.4 &58.2 &60.9 &27.2 &30.2 &41.4 &42.5 &46.8 &34.3 &39.6 &45.1 &48.3 &51.5 \\
				
				&HLML \textit{et al.} \cite{zhang} & FRCN (R50) & 46.8 & 49.2 &50.2 &52.0 &52.4 &39.4 &43.1 &43.6 &44.1 &45.7 &44.1 &49.8 &50.5 &52.3 &52.8   \\

				&HLML \textit{et al.} \cite{zhang} & FRCN (R101) & 48.6 & 51.1 &52.0 &53.7 &54.3 & 41.6 & 45.4 &45.8 &46.3 &48.0 &46.1 &51.7 &52.6 &54.1 &55.0   \\

				& KFSOD \cite{KFSOD} & FRCN (R101)   & 44.6  &  -  & 54.4   & 60.9   & 65.8   & 37.8  &  -  & 43.1   & 48.1   & 50.4   & 34.8   & - & 44.1   & 52.7   & 53.9 \\
				
				& TENET \cite{eccv2022TENET}  & FRCN (R101)  & 46.7  & 52.3  & 55.4  & 62.3  & 66.9  & 40.3  & 41.2  & 44.7  & 49.3  & 52.1  & 35.5  & 41.5  & 46.0  & 54.4  & 54.6 \\
				
				& TENET \cite{eccv2022TENET} &  FRCN (R50)  & 48.5  & 55.2  & 58.7  & 65.8  & \textbf{69.0}  & 42.6  & 43.4  & 47.9  & 52.0  & \textbf{54.2}  & 37.9  & 43.6  & 48.8  & 56.9  & 57.6 \\
				
				&Meta-DETR~\cite{metadetr} & D-DETR (R101) &35.1 &49.0 &53.2 &57.4 &62.0 &27.9 &32.3 &38.4 &43.2 &51.8 &34.9 &41.8 &47.1 &54.1 &58.2 \\
				
				& VFA  \cite{vfa} & FRCN (R101)  & 57.7  & 64.6  & 64.7  & 67.2  & 67.4  & 41.4  & 46.2  & 51.1  & 51.8  & 51.6  & 48.9  & 54.8  & 56.6  & 59.0  & 58.9 \\
				
				& Meta-Tuning \cite{cvpr2023} & FRCN (R101) & 58.4  & 62.4  & 63.2  & 67.6  & 67.7  & 34.0  & 43.1  & 51.0  & 53.6  & 54.0  & 55.1  & 56.6  & 57.3  & 62.6  & 63.7 \\

				\midrule

				& \textit{Single Task} \\	
				
				&	\multicolumn{1}{l|}{LSTD~\cite{lstd}}      & FRCN (R101)            &  8.2                       & 1.0                       & 12.4                      & 29.1 & 38.5 & 11.4                      & 3.8                       & 5.0                       & 15.7 & 31.0 & 12.6                      & 8.5                       & 15.0                      & 27.3                      & 36.3                      \\ 
				
				& MPSR \cite{MPSR} & FRCN (R101) & 41.7 & - & 51.4 & 55.2 & 61.8 & 24.4 & - & 39.2&   39.9 & 47.8 & 35.6 & - & 42.3 & 48.0 & 49.7 \\
				
				&	\multicolumn{1}{l|}{TFA w/ cos~\cite{TFA}}       & FRCN (R101)                       &  39.8                      & 36.1                      & 44.7                      & 55.7 & 56.0 & 23.5                      & 26.9                      & 34.1                      & 35.1 & 39.1 & 30.8                      & 34.8                      & 42.8                      & 49.5                      & 49.8                      \\

				&	\multicolumn{1}{l|}{FSCE~\cite{fsce}}      & FRCN (R101)                       &  44.2                      & 43.8                      & 51.4                     & 61.9 & 63.4 & 27.3                      & 29.5                      & 43.5                      & 44.2 & 50.2 & 37.2                      & 41.9                      & 47.5                      & 54.6                      & 58.5                      \\ 
				
				& SRR-FSD \cite{SRR} & FRCN (R101)   & 46.3  &  51.1  &  52.6  &  56.2  &  57.3  &  31.0  &  29.9  &  34.7  &  37.3  &  41.7  &  39.2  &  40.5  &  39.7  &  42.2  &  45.2 \\
				
				& FSOD-SR \cite {SR} & FRCN (R50)   & 50.1  & 54.4  & 56.2  & 60.0  & 62.4  & 29.5  & 39.9  & 43.5  & 44.6  & 48.1  & 43.6  & 46.6  & 53.4  & 53.4  & 59.5 \\

				& Halluc \cite{hall}  & FRCN (R101)  & 47.0  & 44.9  & 46.5  & 54.7  & 54.7  & 26.3  & 31.8  & 37.4  & 37.4  & 41.2  & 40.4  & 42.1  & 43.3  & 51.4  & 49.6 \\
				
				&Retentive \cite{retina} & FRCN (R101)  & 42.4  & 45.8  & 45.9  & 53.7  & 56.1  & 21.7  & 27.8  & 35.2 &  37.0  & 40.3 &  30.2 &  37.6 &  43.0 &  49.7 & 50.1 \\
				
				& CGDP \cite{cgdp} & FRCN (R101)  & 40.7  & 45.1  & 46.5  & 57.4  & 62.4  & 27.3  & 31.4  & 40.8  & 42.7  & 46.3  & 31.2  & 36.4  & 43.7  & 50.1  & 55.6 \\
				
				& DeFRCN \cite{defrcn}   &FRCN (R101) & 40.2 & 53.6 & 58.2 & 63.6 & 66.5 & 29.5 & 39.7 & 43.4 & 48.1 & 52.8 & 35.0 & 38.3 & 52.9 & 57.7 & 60.8 \\
				
				& FADI \cite{neurips2021}  & FRCN (R101)  & 50.3  & 54.8 &  54.2  & 59.3 &  63.2 &  30.6  & 35.0  & 40.3  & 42.8  & 48.0  & 45.7  & 49.7  & 49.1  & 55.0  & 59.6 \\
				
				& Pseudo-Lb \cite{cvpr2022} & FRCN (R50)    & 50.5   & 53.1   & 56.4   & 61.7   & 62.7   & 36.4   & 33.8  &  46.1   & 49.3   & 48.2   & 42.4   & 44.3   & 49.1   & 55.2   & 57.6 \\
				
				& Pseudo-Lb \cite{cvpr2022}   & FRCN (R101)   & 54.5   & 53.2   & 58.8   & 63.2   & 65.7   & 32.8   & 29.2   & 50.7   & 49.8   & 50.6   & 48.4   & 52.7   & 55.0   & 59.6   & 59.6 \\

				&  CoCo-RCNN  \cite{eccv2022} &  FRCN (R101)    &  33.5   &  44.2   &  50.2   &  57.5   &  63.3   &  25.3   &  31.0   &  39.6   &  43.8   &  50.1   &  24.8  &   36.9  &   42.8  &   50.8  &   57.7 \\
				
				
				& MF-Distillation \cite{eccv2022multifacedistill}  & FRCN (R101)  & \textbf{63.4}  & \textbf{66.3}  & 67.7  & \textbf{69.4}  & 68.1  & 42.1  & 46.5  & 53.4  & \textbf{55.3}  & 53.8  & 56.1  & 58.3  & 59.0  & 62.2  & 63.7 \\
				
				&  KD-BVWR \cite{eccv2022kdisll} &   FRCN (R101) &   58.2  &  62.5  &  65.1  &  68.2  &  67.4  &  37.6  &  45.6  &  52.0  &  54.6  &  53.2  &  53.8  &  57.7  &  58.0  &  62.4  &  62.2 \\
				
				& Calibrations \cite{eccv2022Calibration} & FRCN (R50)  & 40.1  & 44.2  & 51.2  & 62.0  & 63.0  & 33.3  & 33.1  & 42.3  & 46.3  & 52.3  & 36.1  & 43.1  & 43.5  & 52.0  & 56.0 \\
				
				& ECEA \cite{ecea} & FRCN (R101)  & 59.7  & 60.7  & 63.3  & 64.1  & 64.7  & \textbf{43.1} &  45.2  & 49.4  & 50.2  & 51.7  & 52.3  & 54.7  & 58.7  & 59.8  & 61.5 \\

				& VAE-CLIP \cite{localizationcvpr2023} & FRCN (R101)  &  62.1  & 64.9  & \textbf{67.8}  & 69.2  & 67.5  & 39.9  & \textbf{46.8}  & \textbf{54.4}  & 54.2  & 53.6  & \textbf{58.2}  & \textbf{60.3}  & \textbf{61.0}  & \textbf{64.0}  & \textbf{65.5} \\
				
				\bottomrule





				

				
		\end{tabular} }
		
		\label{voc}
	\end{table*}

	\begin{table*}[t]
		\centering	
		\tabcolsep=0.06cm
		
		\caption{G-FSOD performance on three PASCAL VOC all classes (AP50). 
		}
		\scalebox{0.65}{
			\begin{tabular}{@{}l|c|ccccc|ccccc|ccccc@{}}
				\toprule
				\multirow{2}{*}{Methods} 	& \multirow{2}{*}{Framework} &   \multicolumn{5}{c|}{All Split1}                                                                                    & \multicolumn{5}{c|}{All Split2}                                                                                    & \multicolumn{5}{c}{All Split3}                                                                                                         \\ 
				
				&	&                   \multicolumn{1}{c}{K=1}                         & \multicolumn{1}{c}{2}                         & \multicolumn{1}{c}{3}                         & \multicolumn{1}{c}{5}    & \multicolumn{1}{c|}{10}    & \multicolumn{1}{c}{1}                         & \multicolumn{1}{c}{2}                         & \multicolumn{1}{c}{3}                         & \multicolumn{1}{c}{5}    & \multicolumn{1}{c|}{10}     & \multicolumn{1}{c}{1}                         & \multicolumn{1}{c}{2}                         & \multicolumn{1}{c}{3}                         & \multicolumn{1}{c}{5}    & \multicolumn{1}{c}{10}                         \\ 
				\midrule
				
				Retentive \cite{retina} & FRCN (R101)  &  71.3  &  72.3  &  72.1  & 74.0  & 74.6  &  66.8  & 68.4  & 70.2  & 70.7 &  71.5 &  69.0  & 70.9  & 72.3  & 73.9  & 74.1 \\
				
				CFA w/ fc \cite{cfa} &  FRCN (R101)  &  70.3  &  69.5  & 71.0 &  74.4 &  74.9 &  67.0 &  68.0 &  70.2 &  70.8  & 71.5 &  69.1 &  70.1 &  71.6 &  73.3 &  74.7\\

				CFA w/ cos \cite{cfa} & FRCN (R101)  &  71.4  & 71.8  & 73.3  & 74.9  & 75.0  & 66.8 &  68.4  & 70.4  & 71.1  & 71.9  & 69.7  & 71.2  & 72.6  & 74.0  & 74.7 \\
				
				CFA-DeFRCN \cite{cfa} & FRCN (R101)  &  75.0  &  76.0  &  76.8  &  77.3  &  77.3 &   70.4  &  \textbf{72.7}  &  73.7  &  74.7  &  74.2  &  74.7  &  75.5  &  75.0  &  76.2  &  76.6 \\
				
				DiGeo \cite{cvpr2023gfsod} & FRCN (R101)  &  69.7 &   70.6 &   72.4  &  75.4  &  76.1 &   67.5 &   68.4 &   71.4 &   71.6 &   73.6 &   68.6 &   70.9 &   72.9 &   74.4 &   75.0\\

				NIFF-DeFRCN \cite{niff} & FRCN (R101) &   \textbf{75.9}  &  \textbf{76.9}  &  \textbf{77.3}  &  \textbf{77.9}  &  \textbf{77.5}  &  \textbf{70.6} &   71.6 &   \textbf{74.5} &   \textbf{75.1} &   \textbf{74.5} &   \textbf{74.7} &   \textbf{76.0} &   \textbf{76.1} &   \textbf{76.8} &   \textbf{76.7} \\
				\midrule
				
		\end{tabular} }
		\label{allvoc}
	\end{table*}

	\section{Comparison of two Taxonomy FSOD Methods}
	
	In this section, we analyze the classification schemes through theory and experiment, respectively, to summarize the corresponding advantages and disadvantages.
	
	\subsection{Theoretical Analysis}
	
	Episode-task-based FSOD methods leverage meta-learning techniques to quickly adapt the model's parameters or update its architecture to effectively handle unseen object classes. This enables efficient learning from data scarcity examples and the merit of such methods can be summarized as follows:
	
	

	\textbf{Strong adaptive ability without fine-tuning.} Episode tasks assist the FSOD model in learning to learn, which acquires meta-knowledge that enables them to rapidly adapt to new object classes and makes them suitable for fast deployment in real-world scenarios. For example, Meta-RCNN \cite{metarcnn} achieves excellent detection performance by combining the Faster R-CNN with meta-learning to rapidly adapt to a small number of labeled samples. Sometimes, these methods have strong adaptive ability without fine-tuning, e.g., attention-RPN \cite{fan} for enhancing the quality of region proposals and AirDet for autonomous exploration of low-power robots \cite{eccv2022rebot}. AirDet \cite{eccv2022rebot} is constantly learning small but valuable bits of information about query images in episode tasks to automatically adapt to robots’ exploration tasks without fine-tuning. 	
	
	\textcolor{blue}{ \textbf{Mitigating domain shift.} By simulating multiple different combinations of support and query domains in the episode-training phase, the model learns to quickly adapt to new domains. Consequently, when confronted with a new target domain, the model can effectively capitalize on its pre-existing knowledge and strategies to rapidly adjust to the new data distribution, thereby diminishing the effects of domain shift \cite{vfa}. On the other words, the model learns to identify and utilize features relevant to the novel task, rather than solely relying on features from the base domain. This enables the model to better adapt to the data characteristics of the target domain and improve performance in the new domain.}
	

	\textcolor{blue}{ Nonetheless, episode-task-based FSOD methods undergo poor interpretability of what they learned in the novel stage \cite{cvpr2023}. In addition, meta-training requires additional computational resources and time to learn the meta-knowledge and adapt the model to new tasks, making them computationally expensive. Fortunately, single-task-based FSOD can quickly adapt to novel classes without considering the combination of meta-learning and detector. Therefore, the merits of these FSOD methods can be summarized as follows: }
	
	\textbf{Faster training and deployment.} By initializing the model with pre-trained weights, single-task-based FSOD methods can significantly reduce training time. The model can start from a good initialization point, allowing for faster convergence and enabling quick deployment in real-world scenarios. For example, TFA \cite{TFA} illustrates that keeping the feature extraction part of the model unchanged and only fine-tuning the last layer can improve the detection accuracy over the episode-task-based approach. 
	
	
	\textbf{Strong generalization ability.} Single-task methods can greatly improve the generalization ability of the FSOD model by fine-tuning two-stage models. This is because, by leveraging shared patterns and structures in the data, as well as learning common features and representations from the base stage. The models become better equipped to adapt to variations in the novel detection task. DeFRCN as a classical single-task example improves the performance of FSOD at a new level by decoupling Faster R-CNN. Following this concept, some researchers \cite{cfa,ecea, niff} borrow the idea of decoupling to improve the generalization ability of their methods.
	
	\textcolor{blue}{Even so, compared with episode-task-based FSOD methods, they are more likely to suffer sensitivity to domain shifts \cite{defrcn, vfa}. Due to the different sources and imbalance between base and novel samples in single-task-based learned models, the performance may degrade \cite{defrcn}. Furthermore, there is no uniform fine-tuning strategy, in which different FSOD tasks freeze or unfreeze different detection modules \cite{cfa}. For example, TFA fine-tuning strategy \cite{TFA} is directly used in DeFRCN  \cite{defrcn}, and the results can be wildly skewed.}

	\begin{table*}[t]
		\centering
		\tabcolsep=0.07cm
		
		\caption{
			Performance of FSOD on MS-COCO novel classes. 
		}
		\scalebox{0.7}{
			\begin{tabular}{l|c|cccccc|cccccc}
				
				\midrule
				\multicolumn{1}{c|}{\multirow{2}{*}{Methods}} & \multirow{2}{*}{Framework}  & \multicolumn{6}{c|}{10 shots}            & \multicolumn{6}{c}{30 shots}            \\ 
				
				&	& AP   & AP50 & AP75 & APS & APM  & APL  & AP   & AP50 & AP75 & APS & APM  & APL  \\ \midrule
				\textit{Episode Task}   \\
				MetaDet~\cite{metadet}           &   FRCN (VGG16)          & 7.1  & 14.6 & 6.1  & 1.0 & 4.1  & 12.2 & 11.3 & 21.7 & 8.1  & 1.1 & 6.2  & 17.3 \\
				
				FSDetView \cite{xiao}  &   FRCN (R101)             & 10.3 & 25.1 & 6.1  & 3.5 & 11.3 & 14.6 & 14.2 & 31.4 & 10.3 & 4.7 & 15.0 & 21.5 \\
				
				Meta R-CNN~\cite{metarcnn}   &   FRCN (R101)          & 8.7  & 19.1 & 6.6  & 2.3 & 7.7  & 14.0 & 12.4 & 25.3 & 10.8 & 2.8 & 11.6 & 19.0 \\
				
				DCNet \cite{dcnet} &   FRCN (R101)  &12.8 & 23.4 & 11.2 &4.3 &13.8 &21.0  &18.6 &32.6 &17.5 &6.9 &16.5 & 27.4 \\
				
				CME \cite{cme}  &   FRCN (R101) &15.1 &24.6 &16.4& 4.6 &16.6 &26.0  &16.9 &28.0 &17.8& 4.6 &18.0 &29.2 \\

				HLML \textit{et al.} \cite{zhang} & FRCN (R101) &  13.9 & 29.5 & 11.7 & - & - & - & 7.6 & 15.2 & 19.0 & - & - & - \\
				
				FCT \cite{fct} &  FRCN (R101) & 17.1 &  - & - & - & - & - & 21.4 &  - & - & - & - & - \\
				
				QA-FewDet \cite{qa}  & FRCN (R101) &  11.6    &   23.9   &   9.8  &   -   &  -  &   -   &  16.5   &    31.9   &   15.5   &  -   &  -   &   -\\
				
				Meta-DETR \cite{metadetr} & D-DETR (R101)  & 19.0   & 30.5  &  19.7 &  - &  -  & -    & 22.2  &  35.0 &  22.8 &  - &  -  & - \\
				
				Meta FRCN \cite{metafrcn} & FRCN (R101) & 9.7  & 18.5  & 9.0 &  -  & -  & -  & 10.7  & 19.6  & 10.6 &  -  & -  & - \\
				
				TENET  \cite{eccv2022TENET} &  FRCN (R101) & 19.1  &  27.4  &  19.6   & -   & -  & -   & -  & -  & -  & -  & -  & - \\
				
				VFA \cite{vfa} & FRCN (R101) & 16.2  &- &- &- &- &- & 18.9  &- &- &- &- &- \\
				
				CKPC \cite{tip2}    & FRCN (R101)                & 16.6 & 34.4 & 17.2 & 5.9 & 18.3 & 27.5 & 19.9 & 38.1 & 19.7 & 7.8 & 20.9 & 31.3 \\
				
				Meta-Tuning \cite{cvpr2023} & FRCN (R101) & 18.8 &- &- &- &- &- & 23.4 &- &- &- &- &- \\
				
				\midrule
				
				\textit{Single Task} \\
				FSRW \cite{fsrw}    & FRCN (R101)   & 5.6  & 12.3 & 4.6  & 0.9 & 3.5  & 10.5 & 9.1  & 19.0 & 7.6  & 0.8 & 4.9  & 16.8 \\
				
				MPSR \cite{MPSR}      & FRCN (R101)            & 9.8  & 17.9 & 9.7  & 3.3 & 11.3 & 14.6 & 14.1 & 25.4 & 14.2 & 4.0 & 12.9 & 23.0 \\
				TFA   w/cos~\cite{TFA}    & FRCN (R101)          & 9.1  & 17.1 & 8.8  & 3.7 & 8.0  & 14.3 & 12.1 & 22.0 & 12.0 & 4.7 & 10.8 & 18.6 \\
				FSCE~\cite{fsce}          & FRCN (R101)           & 11.4 & 23.3 & 10.1 & 4.5 & 10.8 & 18.7 & 15.8 & 29.9 & 14.7 & 6.1 & 10.8 & 18.6 \\
				SRR-FSD \cite{SRR}        & FRCN (R101)         & 11.3 & 23.0 & 9.8  & -    &  -    &  -    & 14.7 & 29.2 & 13.5 & -    & -     & -     \\
				
				FSOD-SR & FRCN (R50)  & 11.6  & 21.7  & 10.4  & 4.6  & 10.5  & 17.2 &  15.2 &  27.5 &  14.6 &  6.1  & 14.5  & 24.7 \\
				
				CGDP\cite{cgdp}& FRCN (R101)  &  11.3&  20.3 &  - &  - & -&  - & 15.1 &  29.4 & -&  -&  -&  - \\
				
				Retentive \cite{retina}& FRCN (R101)  & 10.5  &-  &-&  - & -&  - &  13.8 & - & -&  -&  -&  - \\

				
				Pseudo-Lb (R101) \cite{cvpr2022}& FRCN (R101)  &17.8 &30.9 &17.8   &-  &-  &-  & 24.5 &41.1 & 25.0   & -  &- & - \\
				Pseudo-Lb (Swin-S) \cite{cvpr2022}& FRCN (R101)  &19.0 &34.1& 19.0  &-  &-  &-  &\textbf{26.8} & \textbf{45.8} &\textbf{27.5}  &-  &-  &-  \\
				
				FADI \cite{neurips2021} & FRCN (R101)  & 12.2 &- &- &- &- &- & 16.1&- &- &- &- &-  \\
				
				CoCo-RCNN \cite{eccv2022}& FRCN (R101)  &  18.1&  30.4&  18.2  &-& -& -&  20.6 & 33.8 & 21.4& -& -& -\\ 
				
				KD-BVWR \cite{eccv2022kdisll} & FRCN (R101)  &  18.9  &   -   &17.8& -& - & - &  22.6  & -  &22.6& -& -&-\\
				
				MF-Distillation  \cite{eccv2022multifacedistill}& FRCN (R101)   & 19.4   & -   & \textbf{20.2}  &  -  &  -  &  -  &  22.7  &  -   & 23.2  &  -   & -   & - \\

				Norm-VAE \cite{localizationcvpr2023}& FRCN (R101)  & 18.7 & - &17.8 &- &- &- & 22.5&- & 22.4 &- &- &-  \\
				
				PTF \cite{tip1}   & FRCN (R101)                  & 11.7 & 22.6 & 10.9 & 5.1 & 12.2 & 16.5 & 15.7 & 30.4 & 14.4 & 7.3 & 16.3 & 21.4 \\
				PTF+KI \cite{tip1}    & FRCN (R101)              & 13.0 & 24.0 & 12.6 & 6.0 & 13.1 & 18.4 & 16.8 & 30.9 & 16.2 & 8.0 & 17.1 & 23.0 \\

				ECEA \cite{ecea}    & FRCN (R101)            & \textbf{19.6} & \textbf{34.6} & 19.2 & \textbf{8.6} & \textbf{19.6} & \textbf{29.3} & 23.1 & 39.5 & 23.8 & \textbf{9.7} & \textbf{23.1} & \textbf{34.6} \\
				\midrule
		\end{tabular}}
		\label{coco}
	\end{table*}

	\begin{table*}[t]
		\centering
		\tabcolsep=0.07cm
		
		\caption{
			Performance of G-FSOD on MS-COCO all classes. 
		}
		\scalebox{0.85}{
			\begin{tabular}{l|c|ccc|ccc|ccc}
				
				\midrule 
				
				\multicolumn{1}{c|}{\multirow{2}{*}{Methods}} & \multirow{2}{*}{Framework} & \multicolumn{3}{c|}{5 shots}             & \multicolumn{3}{c|}{10 shots}            & \multicolumn{3}{c}{30 shots}            \\ 
				
				&	& AP   & bAP & nAP & AP   & bAP & nAP  & AP   & bAP & nAP   \\ \midrule

				Retentive \cite{retina} & FRCN (R101) 	& 31.5 	& 39.2 	& 8.3 	& 32.1 	& 39.2 	& 10.5 	& 32.9 	& 39.3 	& 13.8 \\
				
				CFA w/ fc \cite{cfa} & FRCN (R101) 	& 31.8 	& \textbf{39.5} 	& 8.8 	& 32.2 	& \textbf{39.5} 	& 10.4 	& 33.2	&  \textbf{39.5}	&  14.3 \\
				CFA w/ cos \cite{cfa} & FRCN (R101) 	& 32.0 	& \textbf{39.5} 	& 9.6 	& 32.4 	& 39.4 	& 11.3 	& 33.4 	& \textbf{39.5} 	& 15.1 \\
				
				CFA-DeFRCN \cite{cfa} & FRCN (R101) 	& 33.0 	& 38.9 	& 15.6 	&\textbf{34.0} 	& 39.0	&  \textbf{18.9} 	& \textbf{34.9}	&  39.0 	& \textbf{22.6} \\
				
				DiGeo \cite{cvpr2023gfsod} & FRCN (R101)	&  -	&  - 	& - 	& 32.0	&  39.2	&  10.3	&  33.1	&  39.4 	& 14.2 \\

				NIFF-DeFRCN \cite{niff} & FRCN (R101) 	& \textbf{33.1}	&  38.9	&  \textbf{15.9}	&  \textbf{34.0}	&  39.0 	& 18.8 	& 34.5 	& 39.0 	& 20.9 \\
				
				\midrule
		\end{tabular}}
		\label{allcoco}
	\end{table*}

	\subsection{Experimental Analysis}
	
	In this subsection, we summarize the results of FSOD evaluation on COCO and VOC. Although the LVIS and FSOD datasets are more numerous than the COCO and VOC categories, they need more GPU resources and training times. This is not friendly to researchers with limited budgets and the latter two datasets are just right. Therefore, COCO and VOC are the most popular validated datasets for FSOD. 
	
	\textbf{Results on PASCAL VOC.} Table \ref{voc} lists the benchmark results for the FSOD methods on the PASCAL VOC dataset. From the table, we report the results of the episode-task-based and single-task-based methods, respectively, and give the results of shots $  K = 1, 2, 3, 5, $ and $ 10 $. Meanwhile, we report the average results of each method across multiple seed training detection networks due to the average results are more convincing \cite{TFA}.   
	
	From the final experimental results of each method, the most recent single-task-based VAE-CLIP \cite{localizationcvpr2023} has obtained the best detection results, and 8 groups have obtained the best results among the 15 results. The 13 groups achieved the highest results of VAE-CLIP compared with the latest episode-task-based Meta-Tuning \cite{cvpr2023} among the 15 groups. This result fully demonstrates the merit of single-task-based methods with a simple but effective performance on VOC. However, this does not mean that the former is better than the latter. VAE-CLIP and Meta-Tuning solve different problems, respectively. The VOC dataset is more inclined to the phenomenon that different proposal boxes in the same class are classified into other classes as VAE proposed. In addition, the results of the two categories of VFA \cite{vfa} and ECEA \cite{ecea} methods proposed in the same year are similar. Therefore, in general, both single-task-based and episode-task-based methods are promising research for FSOD.

	Furthermore, in Table \ref{allvoc}, we give the all set results (both of base and novel classes AP50) to report the tolerance of the few-shot model for catastrophic forgetting of base classes with G-FSOD setting. The listed methods in the table follow the single-task-based training strategy to keep the accuracy of base classes. From the table, NIFF-DeFRCN \cite{niff} based on the research of decoupling detection network \cite{defrcn} achieves the best performance. These results can fully verify that such single-task-based models have strong generalization ability. In particular, NIFF \cite{niff} can train a few-shot detector without using the base class during the novel phase, and still ensure that the characteristics of the base class are not forgotten.
	
	\textbf{Results on MS COCO.} The COCO dataset contains 20 novel categories and is more convincing in FSOD model testing than the VOC. Table \ref{coco} shows the results of the COCO dataset for the number of categories $ K = 10 $ and $ 30 $. As can be seen from the table, the test results of Norm-VAE \cite{localizationcvpr2023} and Meta-Tuning \cite{cvpr2023} are similar, and even Meta-Tuning's AP results are higher than the former. Therefore, a single dataset cannot evaluate the performance of models. Furthermore, the results of Table \ref{coco} show that ECEA \cite{ecea} and Pseudo-Lb (Swin-S) \cite{cvpr2022} achieved the best results at 10 and 30 shots, respectively. Both methods consider the accuracy of object localization. ECEA considers extending the obscured features of the object to ensure that the object can be fully predicted, while the latter considers improving the detection quality by introducing a large number of high-quality pseudo-annotations. From these results, there is still a certain development space for improving localization quality to improve the performance of FSOD.
	
	In addition, Table \ref{allcoco} lists the all set results to report the tolerance of the few-shot model for catastrophic forgetting of base classes with the G-FSOD setting on the COCO dataset, which is the same as Table \ref{allvoc}. From the table, We report AP of all classes, bAP50 of base classes, and nAP75 of novel classes of $ K= 5, 10, $ and $ 30 $ shots.

	\section{Challenges, Developing Trends, and Application Scenarios}
	
	\textcolor{blue}{In this section, according to the above theoretical and experimental analysis, we summarize the challenges still existing at current, provide potential research direction and development trends, and briefly introduce the important application fields in FSOD.}
	\subsection{Challenges}
	
	Although FSOD studies have achieved many remarkable results, its detection performance is still far from generic object detection. Furthermore, some challenging tasks need to be further solved. These challenges can be summarized as follows.
	
	\textbf{Limited labeled data.} The primary challenge in FSOD is the scarcity of labeled data for novel or unseen object classes, which constraints make it difficult for models to generalize well to unseen classes. This primary concern in FSOD is the first problem that has been extended frustrating problems, e.g., the similarity of the object is difficult to distinguish \cite{fsce,vfa}, the object has too few utilized features to recognize \cite{fct,metadetr}, etc. When all the extended problems can be solved, i.e., the detection model is trained in large-scale labeled datasets, the few-shot challenge of object detection is no longer a problem.
	
	\textbf{Similar semantic classes.} FSOD models need to understand and generalize visual concepts across different object categories to distinguish subtle differences between similar object classes. However, there can be tiny variations in appearance, shape, and context between the different closely related object categories. The complexity of capturing intricate visual features and limited training examples may confusingly classify the objects with the fine-grained semantic gap. For instance, current researchers \cite{fsce} have suggested a solution to try to solve this question, i.e., the contrastive learning-based approach, but this method does not work in the number of categories of $ K = 1 $ or $ 2 $ or even harms the final prediction. In addition, KG-based methods utilize the semantic relation \cite{SRR,SR} of different objects, but it is difficult to build relation reasoning with semantic relationships on the graph when there is only one object in an image with a popular background. Therefore, bridging this semantic gap between the different classes is still a challenge that affects the model's ability to recognize and detect novel objects accurately.
	
	\textbf{Overfitting and poor generalization.} Overfitting is yet a common issue in few-shot detectors, where models tend to perform well on the base training set but fail to generalize to novel examples. Decoupling the classic object detectors can effectively assist the base models to generalize few-shot detectors and prevent the novel model from overfitting. In recent, both episode-task-based \cite{vfa} and single-task-based \cite{cfa} methods have borrowed the decoupling-based FSOD detector to avoid overfitting and enhance prediction accuracy. According to recent studies \cite{cvpr2023,localizationcvpr2023}, achieving good generalization from base to novel in FSOD is still a challenge due to the lack of training data and the need to extract meaningful and transferable representations from limited examples.
	
	\textbf{Class imbalance and domain shift.} \textcolor{blue}{In FSOD, the number of training examples per class in the novel stage is extremely limited compared to the base. This class imbalance problem can lead to biased model training, i.e. if classification decision features of the novel object are similar to the base object, the model may be inclined to recognize the novel class as the base one according to prior knowledge.} Furthermore, in real-world scenarios, the distribution of base and novel data varies across different domains, making it challenging for FSOD models to generalize well. \cite{vfa} found this problem and proposed VFA to solve this problem. VFA assumes that the distribution of features within a class can be effectively captured by a parametric probabilistic model like a VAE \cite{vae}. However, this assumption may not always hold, especially for complex or highly varied class distributions. If the learned distribution is not expressive enough, it may lead to suboptimal feature aggregation and hinder the performance of the few-shot learning or object detection system. Adapting models to handle domain shifts and transfer knowledge from base to novel is a significant challenge in FSOD.
	
	\textbf{Large-scale data.} The utilization of large models, such as the ImageNet pre-trained model (ResNet-101) \cite{res101}, DALL-E model \cite{dall}, CLIP model \cite{clip}, etc., has shown the potential to enhance the performance of few-shot object detection (FSOD). These large models, with their extensive parameters and deep network structures, can learn more expressive and discriminative feature representations. Consequently, FSOD models can extract more accurate features from limited samples. For instance, in \cite{localizationcvpr2023}, CLIP features are employed in the VAE model to improve FSOD performance. Furthermore, \cite{eccvbase} utilizes novel features from the base set to augment the novel detector. However, in practical applications, if the base data contains novel categories that are known, we label such novel objects to directly train a general object detection detector \cite{tipob1,tipob2,flob} can achieve better results than \cite{eccvbase}. According to the definition of FSOD and practical considerations, the base set should not include novel categories, but some features of base classes that are similar to novel classes can be used to generalize the novel model to enhance novel object representation. Therefore, how to properly use the features provided by large-scale data in unseen classes is still a challenge.
	
	Addressing these challenges requires innovations in algorithms, dataset creation, and training strategies to improve the performance and robustness of FSOD models. Researchers are actively exploring these areas to advance the state-of-the-art in few-shot object detection and make it more applicable in real-world scenarios.

	\subsection{Potential Research Direction and Development Trend}
	
	The future of FSOD holds several potential areas of research and development. As the field progresses, advancements in algorithms, models, and techniques will continue to address the challenges of FSOD, enabling models to learn and adapt effectively with limited labeled data and achieve improved performance.
	
	\textbf{Hybrid approaches.} Hybrid approaches leverage knowledge from related tasks or domains \cite{setnms,crowdob,multicutmultimix} to enhance FSOD performance, e.g., fine-grained learning for FSOD \cite{fgl1,fgl2,fgl3}. Fine-grained objects often contain subcategories with subtle differences, such as different breeds of animals, various plants, specific models of products, etc., which cannot be accurately distinguished by traditional classification methods \cite{fgl4,fgl5}. With the continuous development of fine-grained learning, models have been able to accurately identify objects with tiny differences between these sub-categories. \cite{fgob} shows that the fine-grained classification head can assist the object detection task to achieve exciting results. However, few studies adopt fine-grained learning in the two-stage training paradigm of FSOD. Thus, the application prospect of fine-grained learning in FSOD is very potential. Developing a combination scheme of fine-grained and FSOD models can effectively solve the challenge brought by different classes with similar semantic features in FSOD. In addition, \cite{survey2} considers that the weakly supervised setting \cite{weakly1,weakly2,weakly3,weakly4} and self-supervised learning \cite{self1,self2} can also be used as technical extensions of FSOD in the future.

	\textbf{The quality of localization.} Due to the powerful generation of candidate boxes in the two-stage Faster RCNN detector, the regression task often performs better than the classification task in FSOD. Therefore, most of the FSOD methods mainly focus on object classification and less work has been considered on localization. However, the performance of object localization is still a topic worthy of attention. For example, the RPN module generates several candidate boxes with high confidence in the same object, but one of the candidate boxes of the object could be classified as another category due to the localization quality of the prediction box \cite{localizationcvpr2023}. Candidate boxes of different qualities from the same object may contain different scales, object features, and backgrounds, which could confuse the final classifier. In addition, \cite{ecea} finds that due to the few-shot samples, the novel set may miss part features of the object, resulting in the novel model cannot completely predict the whole object localization. Potential problems based on localization seriously harm the detection accuracy of the FSOD model. Therefore, it is necessary to further explore the FSOD positioning problem in the future for the long-term development of FSOD.
	
	
	\textcolor{blue}{\subsection{The Application of FSOD Scenarios}}
	
	\textcolor{blue}{There are many significant importance and practical application values in several fields that require FSOD algorithms, which can be introduced as follows. }
	
	\textcolor{blue}{\textbf{Traffic safety and accident prevention.} Traffic safety systems need to identify potential hazards on the road, such as wrong-way driving or objects obstructing traffic. It is not practical to label a large number of obstacles that may randomly appear on the road \cite{inllfsod}. With limited labeled data, the FSOD-based model can aid traffic safety systems to accurately identify and monitor different objects in real-time, thereby enabling effective traffic surveillance and management \cite{metalearner3,trafficnew}.}
	
	\textcolor{blue}{\textbf{Medical detection.} Some medical conditions or rare diseases may have limited available data for training specific detection models. FSOD algorithms enable medical models that can identify and detect these rare conditions with only a few labeled examples, aiding in early diagnosis and appropriate treatment planning. For example, \cite{medical1} and \cite{medical2} adopted the episode-task-based algorithm to accurately recognize the few-shot object in medical images. Industrial defect detection. }
	
	\textcolor{blue}{\textbf{Industrial defect detection.} In industrial settings, the need for quality control and defect detection is paramount \cite{agde1}. However, new types of defects or anomalies may emerge as manufacturing processes evolve or new products are introduced. Annotated them for training defect detection models in industrial settings is often scarce and time-consuming to collect. Therefore, FSOD methods can be employed to quickly train industrial defect detection models for identifying specific defects or anomalies in products, minimizing production errors, and ensuring product quality \cite{fault1}.}
	
	\textcolor{blue}{\textbf{Earth Surface Detection.} Identifying earth surface objects in satellite remote sensing imagery is important for urban planning and infrastructure development, disaster management, environmental monitoring, etc \cite{ende2}. Earth surface conditions can change rapidly due to various factors, e.g., natural disasters, urban development, or environmental changes \cite{remotefsod1}. In such scenarios, traditional detection methods that rely on extensive pre-training or re-training may not be feasible. FSOD as a reasonable method can effectively solve such problems \cite{remotefsod3}.  }

	\textcolor{blue}{These research domains are only a part of FSOD which has potential application value in various fields and provides a solution to quickly adapt to unseen object categories and reduce labeling costs \cite{survey3}. With the development of FSOD technology and the deepening of application research \cite{agde1,agde2}, FSOD will play a greater role in more fields.}

	\section{Conclusion}
	
	In this survey, we have provided an overview of the current state-of-the-art approaches and techniques in FSOD. We have categorized the approaches into episode-task-based and single-task-based methods according to the definition of transfer learning. Based on this taxonomy, we then surveyed the latest FSOD algorithms to facilitate a deeper understanding of the FSOD problem and the development of innovative solutions. Finally, we discussed the advantages and limitations of these FSOD methods to overview challenges and future research directions for further advanced FSOD.
	
	\section*{Acknowledgements}
	
	This work is partially supported by the National Key R\&D Program of China under Grant 2022YFC3301704, Cooperation Project of Industry, Education, and Research of Zhuhai under Grant ZH22017001210089PWC, NSFC under Grant 61772220, Special projects for technological innovation in Hubei Province under Grant 2018ACA135, and Key R\&D Plan of Hubei Province under Grant 2020BAB027.

	\bibliographystyle{elsarticle-num}
	\bibliography{PAM}
	
\end{document}